\documentclass[referee,sn-mathphys-num]{sn-jnl}

\usepackage{graphicx}%
\usepackage{multirow}%
\usepackage{amsmath,amssymb,amsfonts}%
\usepackage{amsthm}%
\usepackage{mathrsfs}%
\usepackage[title]{appendix}%
\usepackage{xcolor}%
\usepackage{textcomp}%
\usepackage{manyfoot}%
\usepackage{booktabs}%
\usepackage{listings}%

\usepackage{comment}
\usepackage{algorithmic}
\usepackage{multirow}
\usepackage{subcaption}
\usepackage{makecell}
\usepackage{threeparttable}
\usepackage{url}

\begin{document}

\title{Experimental evaluation of offline reinforcement learning for HVAC control in buildings}


\author[1]{\fnm{Jun} \sur{Wang}}\email{jwang692-c@my.cityu.edu.hk}

\author[1]{\fnm{Linyan} \sur{Li}}\email{linyanli@cityu.edu.hk}

\author[2,3]{\fnm{Qi} \sur{Liu}}\email{qiliuql@ustc.edu.cn}

\author*[1]{\fnm{Yu} \sur{Yang}}\email{yuyang@cityu.edu.hk}

\affil*[1]{\orgdiv{School of Data Science}, \orgname{City University of Hong Kong}, \orgaddress{\country{Hong Kong}}}

\affil[2]{\orgdiv{Anhui Province Key Laboratory of Big Data Analysis and Application}, \orgname{University of Science and Technology of China}, \orgaddress{\city{Hefei}, \state{Anhui}, \country{China}}}

\affil[3]{\orgname{State Key Laboratory of Cognitive Intelligence}, \orgaddress{\city{Hefei}, \state{Anhui}, \country{China}}}



\abstract{Reinforcement learning (RL) techniques have been increasingly investigated for dynamic HVAC control in buildings.
However, most studies focus on exploring solutions in online or off-policy scenarios without discussing in detail the implementation feasibility or effectiveness of dealing with purely offline datasets or trajectories.
The lack of these works limits the real-world deployment of RL-based HVAC controllers, especially considering the abundance of historical data.
To this end, this paper comprehensively evaluates the strengths and limitations of state-of-the-art offline RL algorithms by conducting analytical and numerical studies. 
The analysis is conducted from two perspectives: algorithms and dataset characteristics.
As a prerequisite, the necessity of applying offline RL algorithms is first confirmed in two building environments. The ability of observation history modeling to reduce violations and enhance performance is subsequently studied.
Next, the performance of RL-based controllers under datasets with different qualitative and quantitative conditions is investigated, including constraint satisfaction and power consumption.
Finally, the sensitivity of certain hyperparameters is also evaluated.
The results indicate that datasets of a certain suboptimality level and relatively small scale can be utilized to effectively train a well-performed RL-based HVAC controller.
Specifically, such controllers can reduce at most 28.5\% violation ratios of indoor temperatures and achieve at most 12.1\% power savings compared to the baseline controller.
In summary, this paper presents our well-structured investigations and new findings when applying offline reinforcement learning to building HVAC systems.}




\maketitle

\section{Introduction}

It is estimated that buildings account for approximately 40\% of the global energy consumption~\cite{perez2008review}, where half of it is attributed to the heating, ventilation, and air conditioning (HVAC) systems~\cite{oldewurtel2013importance}.
As a result, it is important to design effective HVAC controllers, which can improve both building energy efficiency and further reduce global carbon emissions while maintaining indoor thermal comfort at the same time.

Currently, there are two main challenges in designing smart HVAC management systems.
First, classic PID-based~\cite{kasahara1999design} or rule-based~\cite{wang2008supervisory} HVAC controllers require a thorough understanding of abstract building models. However, this procedure relies on domain-specific knowledge from experts, making it infeasible for deployment across different buildings.
In practice, the growing complexities of buildings, as well as varying control variables exacerbate this issue. Therefore, ideal HVAC control strategies might as well be model-free.

Second, stochastic thermodynamics, along with volatile exogenous factors like weather conditions, energy pricing, and occupants' behaviors, introduce high uncertainty into the HVAC systems.
With the help of auxiliary prediction tasks, model predictive control (MPC)~\cite{afram2014theory} methods, to some extent, alleviate the dynamics in decision-making through a constrained optimization problem.
However, apart from the dependency on accurate building models, neglecting long-horizon utilities limits the optimality of resulting MPC strategies.

Fortunately, as machine learning (ML) techniques develop, such a dynamic model-free decision-making process can be realized. The most widely adopted solutions~\cite{yu2021review} leverage reinforcement learning (RL) algorithms, where the control strategies are developed through interactions with simulated building environments.
Beneficial from long-horizon maximization of predefined reward functions, RL-based HVAC controllers take into account future trends while making decisions, thus being able to deal with situations where dramatic fluctuations occur in terms of thermal or exogenous conditions.
For example, it is firmly demonstrated that off-policy model-free algorithms (e.g., SAC, TD3) outperform other methods in effectiveness and data-efficiency~\cite{biemann2021experimental, mahbod2022energy, zhang2023deep}.

However, even off-policy solutions still rely on the touch of the building environment during training, at least a simulated one. Few existing studies recognize the significance of HVAC control systems from offline datasets except for \cite{blad2022data, liu2022safe, zhang2022safe}.
Specifically, other than replacing the environment model required by a supervised regression model~\cite{blad2022data}, the remaining studies~\cite{liu2022safe, zhang2022safe} aim to explore environment-free HVAC solutions based on existing offline RL algorithms, such as BCQ and CQL.

However, we still notice the following two challenges. First, as shown in previous methods, considering an auxiliary prediction task, such as indoor environment~\cite{zhu2021fast}, power load~\cite{fan2019cooling, liu2024dynamically}, and occupancy comfort~\cite{dong2018occupancy, zhang2023human}, is beneficial for performing HVAC control.
Essentially, such a prediction module explicitly enables the HVAC controllers to memorize and extract long-term information, which is exactly the defect of current end-to-end RL-based frameworks.
Second, without permission for further data collection during offline training, it is urgent to figure out the appropriate properties of the offline datasets that are suitable for learning HVAC controllers. Unfortunately, research on this line is extremely scarce.
 
Consequently, based on a novel observation history module, this paper systematically evaluates several state-of-the-art off-policy and offline RL algorithms for continuous control, with an emphasis on the intrinsic properties of datasets.
The assessment metrics include energy savings, thermal violations, and utility gains.
In summary, our contributions are listed as follows.
\begin{itemize}
    \item A practical framework for end-to-end model-free offline reinforcement learning is proposed to solve the potential drawbacks of traditional off-policy algorithms. In particular, we first introduce the observation history modeling in RL-based HVAC control algorithms to obtain long-term information extractions. With the strong ability of sequential modeling, the dependence on the auxiliary prediction is removed. Meanwhile, the stability of thermal comfort is enhanced both for off-policy and offline algorithm-based controllers.
    
    \item We first identify two critical characteristics of offline datasets, i.e., quality and quantity. Typically, the concept of \textbf{regret ratio} $\delta_{\tau}$ is introduced to measure the optimality of fixed-horizon trajectories. Furthermore, we describe the quality of datasets by the distribution of $\delta_{\tau}$.
    To empirically generate datasets associated with different levels of $\delta_{\tau}$, Gaussian noises are added uniformly with an exploration probability $\epsilon$. Their relations are empirically found to be near-linear.
    
    \item A holistic investigation in terms of the proposed characteristics of offline data for HVAC systems is conducted.
    Due to the difficulty of on-site deployment, we conduct all experiments in simulated building environments like previous studies~\cite{biemann2021experimental, zhang2022safe, mahbod2022energy, he2023predictive} did.
    Our investigation demonstrates that it is intriguingly observed that for offline scenarios, the existence of sub-optimal data is compulsory for training offline RL algorithm-based HVAC controllers.
    In addition, unlike off-policy RL solutions, the quantity requirement for acquiring an equivalent offline HVAC manager is up to a magnitude less than the former, given applicable datasets.

    
\end{itemize}

The rest of the paper is organized as follows.
Section~\ref{sec_literature} briefly introduces related literature on classic and RL-based HVAC controllers.
Section~\ref{sec_preliminary} recalls basic MDP formulation and popular actor-critic framework.
Section~\ref{sec_method} provides theoretical explanations of the necessity and foundations of offline RL algorithms. The incorporation of the observation history modeling module is also included.
Section~\ref{sec_exp_setup} presents the details of the experimental setup.
Section~\ref{sec_exp_results} reports experiments and results with analysis.
Finally, Section~\ref{sec_conclusion} concludes
the paper with a discussion of future directions.
\section{Literature Review}\label{sec_literature}

Roughly, HVAC system control methods can be categorized as classic methods (including rule-based, PID-based and MPC-based ones) and algorithm-based methods. For the latter, we put an emphasis on the reinforcement-learning-based (RL-based) ones in this paper.

\subsection{Classic HVAC Control for Buildings}

    Existing deployed HVAC controllers include simple PID on/off controllers~\cite{kasahara1999design} and rule-based control~\cite{wang2008supervisory}.
    These techniques approach decision-making by sensing the current environment and responding according to multiple pre-defined rules.
    These predefined rules rely heavily on the experiences of experts and trials.
    Typically, such classic methods can not deal with situations where dramatic fluctuations occur in terms of indoor or exogenous conditions. The main reason is that the future trends are not taken into account.
    As a result, a bunch of control approaches based on model predictive control (MPC)~\cite{afram2014theory} have been developed.
    Other than historical data for prediction, domain-specific knowledge is included as well to formulate constrained optimization problems.
    
    However, similar to rule-based methods, detailed and accurate modeling of buildings plays a critical role in the process, which might not be feasible for widespread deployment.
    The performance of MPC-based controllers is limited by the accuracy of the models. Obtaining accurate models is difficult because of the need for extensive intervention by domain experts for every building architecture as well as the difficulties in modeling external factors such as weather, price of electricity, occupancy, IT equipment loads, etc.

\subsection{RL-based HVAC Control for Buildings}
    
    Due to the increasing popularity of machine learning (ML) techniques in the community of HVAC systems~\cite{chen2021data}, control methods based on reinforcement learning (RL) algorithms also develop and flourish~\cite{yu2021review}. Early in the tabular case, researchers have already studied a Q-learning-based controller to manage discrete on/off actions~\cite{henze2003evaluation}.
    However, an inevitable obstacle to RL-based controllers is the demand for data. As discussed in \cite{henze2003evaluation}, significant amounts of data for training are required in order to achieve a performance similar to traditional predictive methods.
    A straightforward solution involves transferring the training process within the simulated environment, beginning with the work of \cite{liu2006experimental}.

    Recently, benefiting from neural network techniques, more end-to-end HVAC control algorithms have been proposed to outperform classical methods. Generally,  these algorithms fall into two categories: learning discrete or continuous control policies.
    In the realm of HVAC control systems for buildings, deep Q-learning (DQN) and its variants, such as double DQN, are preferred for discrete action spaces. Given a certain state, these methods directly represent the potential values for each action by a Q function approximated by a neural network. Consequently, such control methods are shown to be energy-efficient~\cite{gupta2021energy} and defeat rule-based~\cite{wei2017deep} and PID-based~\cite{yuan2021study} controllers.
    \citeauthor{he2023predictive}~\cite{he2023predictive} also integrate the MPC module into the Deep RL-based control framework to further enhance the performance.
    However, such discrete models might encounter the curse of dimensionality when applied to continuous or multivariate control settings~\cite{wei2019deep}. To mitigate this issue, branching strategies~\cite{lei2022practical} and clustering-based control policies~\cite{homod2023deep, he2023efficient} are proposed particularly for high-dimensional discrete HVAC control.

    On the other hand, to directly handle continuous action spaces (e.g., temperature setpoints), the deep deterministic policy gradient (DDPG)~\cite{lillicrap2015continuous} algorithm is a prominent choice and widely applied in \cite{li2019transforming, gao2020deepcomfort, du2021intelligent}. First, compared to discretizing the continuous action spaces and using DQN, these continuous control methods are validated to perform better~\cite{du2021intelligent}.
    Second,  unlike HVAC controllers based on on-policy algorithms~\cite{zhang2019building, azuatalam2020reinforcement}, off-policy methods improve data efficiency and are better suited for real-world applications, particularly with rule-based regularization~\cite{liu2023rule}.
    Furthermore, the subsequent literature~\cite{biemann2021experimental, mahbod2022energy, zhang2023deep} has confirmed that HVAC controllers based on advanced off-policy continuous algorithms, such as Soft Actor-Critic (SAC)~\cite{haarnoja2018soft} and Twin Delayed Deep Deterministic Policy Gradient (TD3)~\cite{fujimoto2018addressing}, outperform others.
    
    In addition, it is worth noting that to reduce the complexity of state or action spaces, designing HVAC controllers based on multi-agent algorithms is a promising direction~\cite{kazmi2019multi, yu2020multi, blad2021multi, blad2023laboratory}, whether for discrete or continuous policies.

    Finally, among those data-driven approaches, \cite{costanzo2016experimental} and \cite{ruelens2016residential} first recognized the importance of offline RL methods. They extended the fitted Q-iteration (FQI) algorithm and incorporated domain knowledge into policy postprocessing to improve performance in discrete on/off control tasks.
    However, their methods are actually value-based ones, where the application to continuous action spaces is difficult. More recently, \cite{blad2022data} proposed to consider a supervised regression model (SRM) to rebuild the environment model from offline data, which essentially belongs to model-based approaches. A major concern with this approach is the model's ability to generalize state-action pairs that deviate significantly from the offline data.
    In contrast, \cite{liu2022safe} and \cite{zhang2022safe} explore model-free methods that directly train algorithm-based controllers without the need for virtual models. To be specific, \citeauthor{liu2022safe}~\cite{liu2022safe} consider a Kullback-Leibler (KL) regularization term based on batch-constrained Q-learning (BCQ) \cite{fujimoto2019off} while \citeauthor{zhang2022safe}~\cite{zhang2022safe} validate the effectiveness of conservative Q-learning (CQL) \cite{kumar2020conservative} as well as model-based policy evaluation for the sake of safe guarantee.

    Along the line of data-driven methods, a compelling yet unexplored issue is how the characteristics of offline data influence the performance of deployed algorithm-based controllers.
    Our paper aims to solve this issue.

\section{Preliminary}\label{sec_preliminary}

This section will first present the basic formulation of reinforcement learning, i.e., the Markov decision process (MDP), followed by a popular algorithmic framework named actor-critic methods.

\subsection{MDP Formulation}
    
    Reinforcement learning (RL) is a computational approach to make decisions within the interactions between the environment and an agent. A basic assumption associated with RL is that all decision goals can be described as the maximization of the expected cumulative reward. As a result, the Markov decision process (MDP)~\cite{sutton2018reinforcement} is proposed to handle the conditions of uncertainty. Formally, MDP represents a quintuple $(\mathcal{S}, \mathcal{A}, P, \rho, r)$ where
    \begin{itemize}
        \item[-] $\mathcal{S} \subset \mathbb{R}^n$ is the state space composed of state $s$.
    
        \item[-] $\mathcal{A}$ is the action space that can be discrete, continuous, or hybrid.  $a \in \mathcal{A}$ is a specific action.
    
        \item[-] $P(s'|s,a)$ is the state transition function, which describes the probability of achieving the next-state $s'$ given the state-action pair $(s,a)$.

        \item[-] $\rho(s)$ is the initial distribution that describes the probability of starting an episode in state $s$.
    
        \item[-] $r(s, a, s')$ is the reward (or negative cost) function to quantify the value of taking action $a$ in state $s$, which might also correspond to the next-state $s'$.
    \end{itemize}
    Finally, the target is to learn a policy $\pi: \mathcal{S} \rightarrow \mathcal{A}$ that maximizes
    \begin{equation}\label{eq_obj}
        J(\pi)=\mathbb{E}_{s_0 \sim \rho, a_t \sim \pi(s_t), s_{t+1} \sim P(s_{t+1}|s_t,a_t)} \sum_{t=0}^T \gamma^t r(s_t, a_t, s_{t+1}).
    \end{equation}
    Here, $\gamma$ is the discount factor for the cumulative reward. Obviously, the distribution of the entire trajectory $\tau=(s_0, a_0, r_0, s_1, \cdots, s_T, a_T, r_T)$ is jointly determined by the initial state distribution $\rho(s_0)$, the state-transition probability $P(s_{t+1}|s_t,a_t)$ and the policy $\pi(s_t)$.

    Concerning the control of HVAC systems, RL can be utilized to represent the interactions between the simulated or true facility environment and algorithm-based controller. Generally, the state space consists of indoor conditions as well as exogenous factors (e.g., weather conditions and human activities), whereas the action space is assumed to be the continuous values for either temperature or flow rate setpoints.
    Since the basic goal of HVAC control is to reduce building energy consumption and improve occupant comfort conditions, the reward function has to consider both factors and further take a balance.
    The detailed design of the reward function for our HVAC control algorithms will be introduced in Section~\ref{sec_exp_rwd}.

\subsection{Actor-Critic Framework}

    Actor-Critic is a framework that combines the advantages of value-based and policy-based methods for solving RL problems. Like policy-based methods, the actor explicitly specifies a parameterized policy $\pi_{\theta}$, which can represent continuous action spaces in a more convenient way. Another critical point is to maintain a parameterized state-action value function $Q_{\phi}(s,a)$, which estimates the expected cumulative reward when taking action $a$ in state $s$.
    The main difference with the value-based methods lies in that the actor is updated by the direction of maximizing the $Q$ values instead of rollout returns or Monte Carlo estimations.
    In other words, the gradient of parameters $\theta$ is
    \begin{equation}\label{eq_actor}
        \nabla_{\theta} J_{\theta}(\pi) = \mathbb{E} _{s \sim \rho^{\pi}(s)} \left[ \nabla_a Q^{\pi}(s,a) \mid _{a=\pi_{\theta}(s)} \nabla _{\theta} \pi_{\theta}(s) \right],
    \end{equation}
    where $\rho^{\pi}(s)$ is the marginal distribution of states induced by the policy $\pi$.
    
    Meanwhile, the learning of the critic itself depends on the recursive iterations from the Bellman equation, which is
    \begin{equation}\label{eq_critic}
        Q^{\pi}\left(s,a \right) = \mathbb{E} _{s' \sim P, a \sim \pi(s)} \left[ r + \gamma Q^{\pi} \left(s', a' \right)  \right].
    \end{equation}

    By convention, the learning of the actor and the critic is performed by the iteration of policy evaluation (Eq.~\eqref{eq_critic}) and policy improvement (Eq.~\eqref{eq_actor}). Here, we use the superscript $\pi$ for $Q^{\pi}$ to stress that a process is on-policy
    if the expectation in Eq.~\eqref{eq_critic} is estimated with transitions that obey $a \sim \pi(s)$. However, in practice, due to the time consumption for convergence and data efficiency, there is another off-policy training manner, which is adopted by recent state-of-the-art RL methods.
    
\begin{figure*}[t]
    \centering
    \begin{subfigure}[h]{0.33\linewidth}
        \includegraphics[width=1.\textwidth]{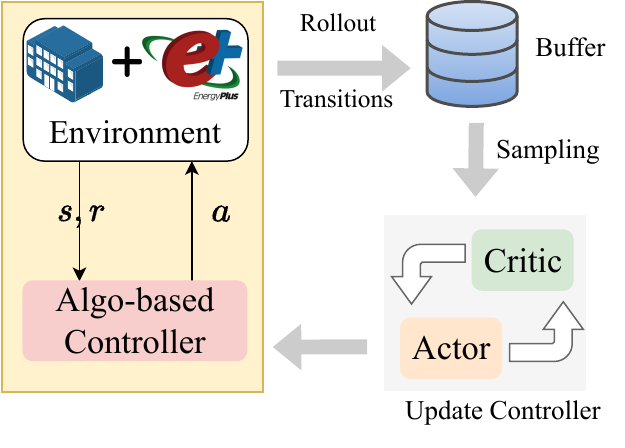}
        \subcaption{}
        \label{fig_flow_offpolicy}
    \end{subfigure}
    \hfill
    \begin{subfigure}[h]{0.325\linewidth}
        \includegraphics[width=1.\textwidth]{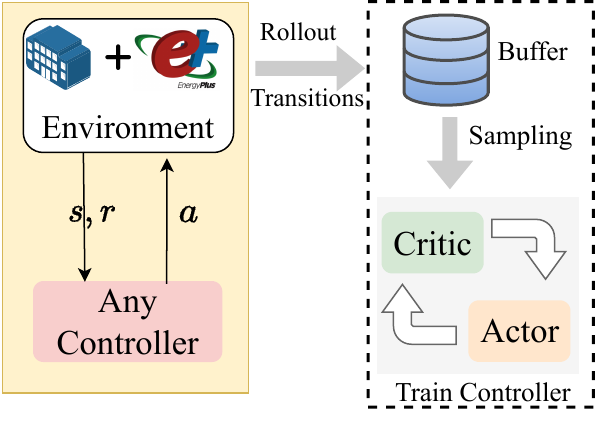}
        \subcaption{}
        \label{fig_flow_offline}
    \end{subfigure}
    \hfill
    \begin{subfigure}[h]{0.305\linewidth}
        \centering
        \includegraphics[width=1.\textwidth]{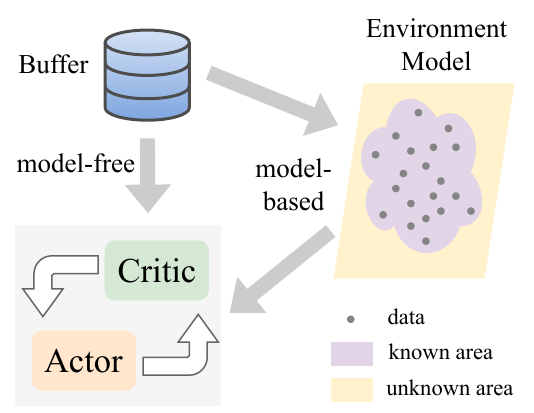}
        \subcaption{}
        \label{fig_flow_classify}
    \end{subfigure}
    \caption{Illustration of (a) the training flow of off-policy RL algorithms, (b) the training flow of offline RL algorithms and (c) the difference between model-free and model-based offline RL algorithms.}
    \label{fig_flow}
\end{figure*}

    As shown in Figure~\ref{fig_flow_offpolicy}, once there exists a replay buffer storing historical transitions, those transitions that do not obey $a \sim \pi(s)$ can be utilized as well to estimate the expectation during the policy evaluation process.
    Next, we introduce two famous off-policy RL algorithms, i.e., twin delayed deep deterministic policy gradient (TD3~\cite{fujimoto2018addressing}) and soft actor-critic (SAC~\cite{haarnoja2018soft}). At a high level, both algorithms are based on the deep deterministic policy gradient (DDPG~\cite{lillicrap2015continuous}), which is designed for deterministic policies.
    To address the issue of overestimation of $Q$ functions, TD3 first proposes three critical tricks and facilitates the training. Meanwhile, SAC forms the bridge between stochastic policy optimization and DDPG-style approaches, by means of considering the entropy regularization.
    As a result, the fundamental objective described in Eq.~\eqref{eq_obj} is changed as
    \begin{equation*}
        J(\pi)=\mathbb{E}_{s_0 \sim \rho, a_t \sim \pi, s_{t+1} \sim P} \left[\sum_{t=0}^T \gamma^t ( r(s_t, a_t, s_{t+1}) + \alpha H(\pi(\cdot \mid s)) ) \right],
    \end{equation*}
    where $H(\pi(\cdot \mid s))$ is the state-conditioned entropy of stochastic policy $\pi$ and $\alpha$ is the trade-off factor.

    Even in the field of HVAC systems, the excellent performance of these off-policy algorithms has been demonstrated~\cite{biemann2021experimental, mahbod2022energy, zhang2023deep}. Hence, TD3 and SAC are selected as our baseline methods for the comparison with the offline algorithms introduced in the next section.

\section{Methodology}\label{sec_method}

In this section, we will introduce the offline RL algorithms, which derive quite different control policies for static datasets compared with vanilla off-policy ones. The divergences will be explained with an emphasis on the extrapolation error issue and regularization-based solutions. Then, we extend the previous offline algorithms with partially observable modeling.

\subsection{Offline RL}

    It should be noted that the interactions with the environment are still required for the training process of off-policy RL algorithms to collect the rollout transitions. We provide the following threefold issues.
    First of all, the demand to interact with real HVAC systems might harm the occupants and be costly. Secondly, despite wide application of simulator-based approaches, the construction of a simulated facility environment requires the experience of experts, especially for new facilities. Thirdly, previously collected interaction data from other controllers (e.g., rule-based controllers) is wasted and not efficiently leveraged.
    Therefore, data-driven methods are critical and promising for optimizing algorithm-based controllers directly with the mixture of offline data.

    Therefore, RL algorithms for offline scenarios are noticed since further interactions with the environment are prohibited, as illustrated in Figure~\ref{fig_flow_offline}. The exclusive training source is the static dataset composed of previously collected trajectories.

    Unfortunately, off-policy methods might not work for offline data. To explain this, we first represent the (mixed) policy that generates the offline data as a behavioral policy $\pi_{\beta}$, which can be estimated through log-likelihood maximization as
    \begin{equation}
        \pi_{\beta}(a \mid s) = N(s, a) / \sum\nolimits_{\tilde{a}} N(s, \tilde{a} ),
    \end{equation}
    where $N(s, \tilde{a} )$ is the frequency the state-action pair $(s, \tilde{a} )$ is observed in offline data.
    
    During off-policy training, the replay buffer, which stores historical transitions and is continuously updated, actually behaves as temporary offline data.
    Although off-policy RL algorithms allow the critic to leverage transitions generated by different policies in the replay buffer, 
    the distance between the target policy during training and the behavioral one generating historical transitions is implicitly restricted~\cite{fujimoto2019off}.
    This is basically caused by the gradual parameter updating in addition to the exploration procedure when computing the target value $r + \gamma Q^{\pi} (s', a')$ in Eq.~\eqref{eq_critic}. In practice, such a strategy often limits the distance of explored data from existing one.
    For example, in DDPG-style algorithms, exploration is realized by adding noises on $\pi(s')$  from Gaussian distributions with limited deviation.


\subsection{Main Challenge}

    When it comes to offline data, such distance between the target policy during optimization and the behavioral one $\pi_{\beta}$ is not guaranteed anymore. This will lead to a severe issue called \textbf{extrapolation error}.
    Intuitively, when the expectation of the target value $r + \gamma Q^{\pi} (s', a')$ is estimated through sampled transitions $\{(s_i,a_i,r_i,s'_i)\}_{i=1}^{N}$, it is most likely to encounter a mismatched distribution of the next state-action pairs $(s', a')$ arising from the discrepancy between $\pi(\cdot|s')$ and $\pi_{\beta}(\cdot|s')$. Consequently, it will induce an inevitable estimation error for the current state-action values $Q(s, a)$ according to Eq.~\eqref{eq_critic}.
    To make it even worse, such estimation error would be accumulated due to the recursive iterations, finally making the Q values fail to converge.
    
    
    In essence, the main reason for out-of-distribution (OOD) mismatches is still the incomplete data coverage of static data and the lack of online correction.
    To empirically exhibit the differences between such two classes of RL algorithms, we design two scenarios with different data collection strategies and analyze the results. Details can be referred to in Section~\ref{sec_exp_rq1}.

    There are two main directions to relieve the distribution shift. As listed in Figure~\ref{fig_flow_classify}, the first one is to reconstruct the environment model through offline data, which develops into model-based approaches~\cite{kidambi2020morel}. For HVAC systems, \citeauthor{blad2022data}~\cite{blad2022data} consider this way and apply a supervised regression model (SRM) as the environment model. So far, finding a well-generalized environment model is still an open question, especially with limited data.
    While the other line is to directly optimize policies without additional models, this paper focuses on these model-free solutions. 

\subsection{Practical Algorithms}

    Given an offline data $\mathcal{D}$, we first specify the iteration process of model-free policy evaluation as
    \small{\begin{equation}\label{eq_policy_eval}
        Q^{k+1} \leftarrow \arg \min _Q \mathop{\mathbb{E}}\limits_{(s, a, s') \sim \mathcal{D}}\left[\left(\left(r(s, a)+\gamma \mathbb{E}_{a' \sim \pi^k \left(\cdot \mid s'\right)} Q^k\left(s', a' \right)\right)-Q^k(s, a)\right)^2\right]
    \end{equation}}
    and the policy improvement procedure as
    \begin{equation}\label{eq_policy_improve}
        \pi^{k+1} \leftarrow \arg \max _\pi \mathbb{E}_{s \sim \mathcal{D}, \bar{a} \sim \pi^k(\cdot \mid s)}\left[ Q^{k+1}(s, \bar{a})\right]
    \end{equation}
    separately. It is confirmed that imposing regularization terms is compulsory~\cite{levine2020offline}.
    Under the actor-critic framework, such a regularization term can be realized for either Eq.~\eqref{eq_policy_eval} or Eq.~\eqref{eq_policy_improve}.

    At the very beginning, methods like BCQ~\cite{fujimoto2019off} or BEAR~\cite{kumar2019stabilizing} adopt the policy constraints. The general idea is concluded as
    \begin{equation}
        \pi^{k+1} \leftarrow \arg \max _\pi \mathbb{E}_{s \sim \mathcal{D}, \bar{a} \sim \pi^k(\cdot \mid s)}\left[ Q^{k+1}(s, \bar{a})\right], \text{ s.t. } D(\pi, \pi_{\beta}) \leq \epsilon.
    \end{equation}\label{eq_constrained_improve}
    In other words, the target policy $\pi$ is forced to avoid deviating far from the behavioral one $\pi_{\theta}$, regardless of explicitly learning a conditioned variational autoencoder (CVAE) (e.g., BCQ) or implicitly estimating maximum mean discrepancy (MMD) (e.g., BEAR). Subsequently, TD3+BC~\cite{fujimoto2021minimalist} tests the effectiveness of a minimalistic form in performing regularization as
    \begin{equation}
        \pi^{k+1} \leftarrow \arg \max _\pi \mathbb{E}_{s,a \sim \mathcal{D}, \hat{a} \sim \pi^k(\cdot \mid s)}\left[ Q^{k+1}(s, \hat{a})+ \lambda \cdot \pi(a \mid s) \right].
    \end{equation}

	Alternatively, a pessimistic value function is taken into account to penalize the uncertain estimation of state-action pairs outside of $\mathcal{D}$. CQL~\cite{kumar2020conservative}, the current state-of-the-art algorithm, just stems from this insight and proposes to lower bound the estimated $Q$ values of state-action pairs $(s,a)$ under the current policy while maximizing the values of those contained in $\mathcal{D}$ at the same time. Therefore, Eq.~\eqref{eq_policy_improve} is modified as
    \small{\begin{equation}
        \begin{aligned}
        Q^{k+1} \leftarrow \arg \min _Q & \left\{ \frac{1}{2} \mathop{\mathbb{E}}\limits_{(s, a, s') \sim \mathcal{D}}\left[\left( \mathcal{B}^{\pi^k}Q^k(s, a)-Q^k(s, a)\right)^2\right] + \right. \\
        & \left. \alpha \cdot \left( \mathbb{E}_{s \sim \mathcal{D}, a \sim \pi^k(\cdot \mid s)} Q^k(s, a) - \mathbb{E}_{s \sim \mathcal{D}, a \sim \pi_{\beta}(\cdot \mid s)}Q^k(s, a) \right) \right\}
    \end{aligned}\end{equation}}
    where $\mathcal{B}^{\pi^k}Q^k(s, a) = r(s, a)+\gamma \mathbb{E}_{a' \sim \pi^k \left(\cdot \mid s'\right)} Q^k\left(s', a' \right)$.

    In the following experiments, we compare the HVAC control performance based on methods from both two directions.

    \begin{figure}[h]
        \centering
        \includegraphics[width=0.75\columnwidth]{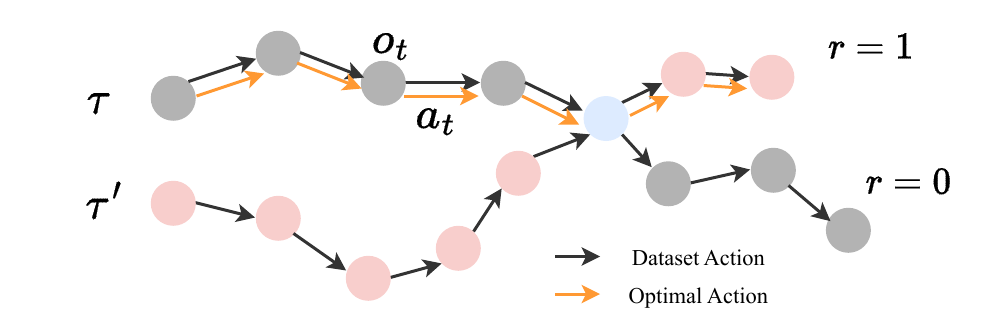}
        \caption{Illustration of modeling the observation history. Once a fully observable state $s_t$ becomes a partially observable observation $o_t$, considering the history sequence $h_t=o_{1:t}$ would be better.}
        \label{fig_obs_history}
    \end{figure}

\subsection{Observation History Modeling}

    In many complicated tasks like the building HVAC control, the state of the whole system can not be completely described and thus is only partially observable in the standard MDP. For example, many HVAC controllers~\cite{costanzo2016experimental, he2023predictive} depend on an auxiliary model to manage a forecast of the exogenous data like weather conditions or thermal data like cooling/heating load.

    The exact theoretical formulation of such cases is the partially observable Markov decision process (POMDP).
    However, the definition of the state-conditioned observation distribution function $\mathcal{E}$ such that $o \sim \mathcal{E}(\cdot \mid s)$ is quite difficult in practice. As a result, a simplified version of observation-history-MDPs~\cite{timmer2009reinforcement} is derived to only replace the state space $\mathcal{S}$ with the observation histories $\mathcal{H}$, as well as adding a horizon $H$.
    
    As shown in Figure~\ref{fig_obs_history}, when action $a \in \mathcal{A}$ is taken given an observation history sequence $h \in \mathcal{H}$, the agent achieves the next sequence as $h'$ through concatenating $h$ and $o' \sim P(\cdot \mid \tau, a)$, followed by the reward $r(\tau, a)$.
    Compared with extra modeling of $\mathcal{E}(\cdot \mid s)$, transforming the one-step policy $\pi(\cdot \mid o)$ to the historical one $\pi(\cdot \mid h)$ helps stabilize the training process. Suppose observation histories can be encoded into latent low-dimensional spaces $\xi: \mathcal{H} \rightarrow \mathcal{Z} \in \mathbb{R}^d$, we can further improve the efficiency of offline RL algorithms~\cite{hong2023offline}. The underlying intuition is to only capture the necessary features for action selection.
    
    One practical way of implicitly learning observation histories representations is to integrate encoders with sequential memorizing ability into the end-to-end offline learning process. We describe one possible structure, i.e, Transformer~\cite{vaswani2017attention} below, which forms a solid basis for the state-of-the-art performance in sequential modeling tasks such as decision making~\cite{chen2021decision} and large language models (LLMs)~\cite{radford2018improving, touvron2023llama}.
    Briefly, after adding position embeddings $\mathbf{p}_{1:t}$ on the features extracted from inputted observation sequences $o_{1:t}$, multiple self-attention blocks are stacked, each with a self-attention (SA) layer and a point-wise feed-forward (FFN) layer in turns. Critical tricks like residual connections and layer normalization ensure the successful training of such a deep structure. The whole encoding process can be concluded as
    \small{\begin{equation}\begin{gathered}
        \mathbf{x}_{(0)} = \mathbf{e} = \text{Features} \left(o_{1:t} \right) + \mathbf{p}_{1:t}, \\
        \Delta\mathbf{x}_{(k)} = \text{SA} \left(\mathbf{x}_{(k-1)}\right) = \text{Attention}\left(\mathbf{x}_{(k-1)} \mathbf{W}^Q_{(k-1)}, \mathbf{x}_{(k-1)} \mathbf{W}^K_{(k-1)}, \mathbf{x}_{(k-1)} \mathbf{W}^V_{(k-1)} \right), \\
        \mathbf{x}_{(k)} = \text{LayerNorm} \left(\mathbf{x}_{(k)} + \Delta \mathbf{x}_{(k)} \right), \\
        \Delta \mathbf{x}_{(k)} = \text{FFN} \left(\mathbf{x}_{(k)} \right) = \text{ReLU} \left( \mathbf{x}_{(k)}\mathbf{W}^{(1)}_{(k)} + \mathbf{b}^{(1)}_{(k)} \right) \mathbf{W}^{(2)}_{(k)} + \mathbf{b}^{(2)}_{(k)}, \\
        \mathbf{x}_{(k)} = \text{LayerNorm} \left(\mathbf{x}_{(k)} + \Delta \mathbf{x}_{(k)} \right), \\
    \end{gathered}\end{equation}}
    where the number of blocks stacked is $k$ and $\mathbf{p}_{1:t}$, $\mathbf{W}^*_{(k)}$ and $\mathbf{b}^*_{(k)}$ are all trainable parameters.

    Note that the representation process can be further enhanced by explicitly introducing objectives like contrastive terms to distinguish different historical sequences. We will discuss this in the future work.
\section{Experiment Setup}\label{sec_exp_setup}

In this section, we use two cases to evaluate offline RL algorithms.
Due to the complex comparison requirements and the difficulty in on-site deployment of various algorithm-based HVAC controllers, we rely on the simulated building models to collect offline data and evaluate model performance, which follows previous studies~\cite{biemann2021experimental, zhang2022safe, mahbod2022energy, he2023predictive}.
After introducing the details of building environments and experimental settings, we list the research questions concerned.

\subsection{Building Environments}
    Two gym~\cite{brockman2016openai}-like building environments are applied to serve as the evaluator and exhibit how the learned control policies perform after deployment. Again, the key training process of offline HVAC controllers does not depend on such simulators.
    So far, there are two main manners to realize runtime HVAC control based on a simulator like EnergyPlus~\cite{crawley2001energyplus}. The first approach is based on the software framework named building controls virtual testbed (BCVTB), whereas the other one relies on a high-level control method, i.e., energy management system (EMS), which uses the EnergyPlus Runtime Language (Erl) as the scripting language.

    \begin{figure}[ht]
        \centering
        \includegraphics[width=0.5\columnwidth]{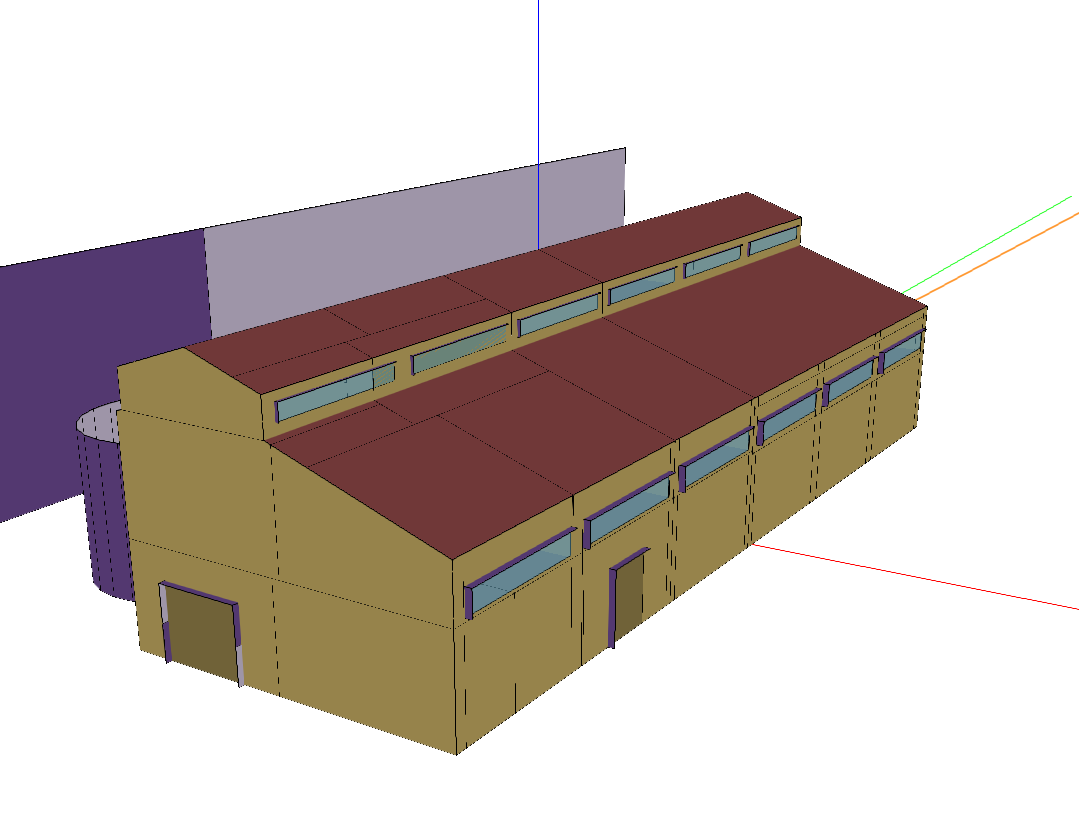}
        \caption{Overview of the MixedUse facility.}
        \label{fig_building_mixeduse}
    \end{figure}
    
    The first case we select is the MixedUse~\footnote{https://bsl546.github.io/energym-pages/sources/mixeduse.html} facility included in the Energym~\cite{scharnhorst2021energym} benchmark, where all building models can be controlled obeying the first manner. Specifically, MixedUse is a building with 13 thermal zones (with 8 controllable) used for residential and office spaces. The overview is shown in Figure~\ref{fig_building_mixeduse}. The HVAC system installed consists of two Air Handling Units (AHU), one dedicated exclusively to thermal zones 5, 6 and 7 (only zone 5 controllable) and a second one for remaining thermal zones.
    As a result, all controllable variables are the thermostat setpoint of 8 zones, as well as the fan flow setpoint and temperature setpoint of two AHUs.

    \begin{figure}[ht]
        \centering
        \includegraphics[width=0.78\columnwidth]{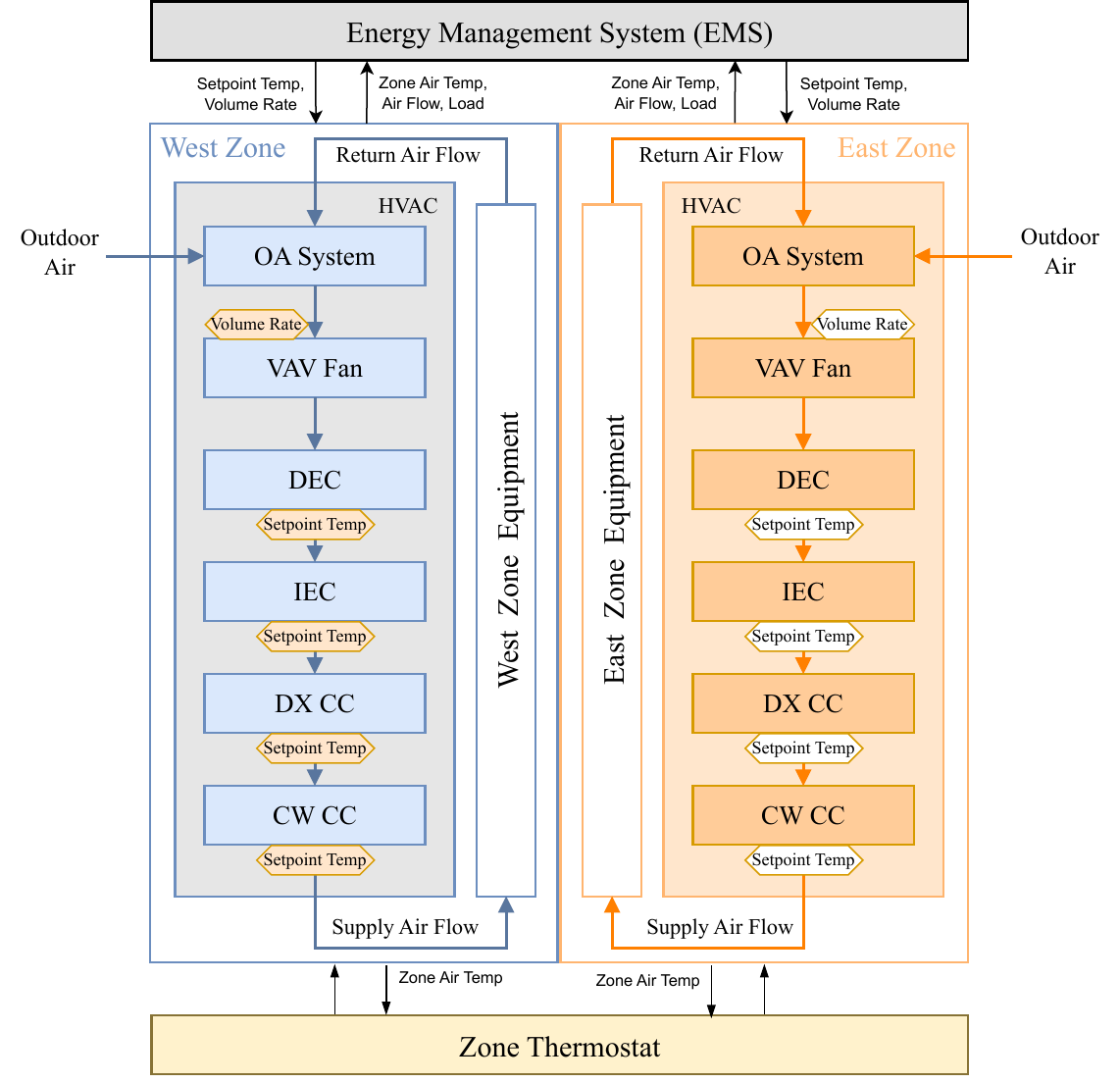}
        \caption{Control flow of the 2-Zone DataCenter facility.}
        \label{fig_building_datacenter}
    \end{figure}
    
    The second one is a two-zone DataCenter facility reformed by \citet{moriyama2018reinforcement} under the second control manner mentioned above. Each zone has a dedicated HVAC system, which is connected through a ``supply air'' duct and a ``return air'' duct to exchange heat. As illustrated in Figure~\ref{fig_building_datacenter}, each HVAC system is composed of several components connected sequentially, including the outdoor air system (OA System), variable volume fan (VAV Fan), direct evaporative cooler (DEC), indirect evaporative cooler (IEC), direct expansion cooling coil (DX CC), and chilled water cooling coil (CW CC). Among them, DEC, IEC, DX CC, and CW CC are controlled by a single shared setpoint temperature. In addition, the air volume supplied to each zone adjusted by the VAV Fan is also controllable.

\begin{table}[ht]
    \centering\small
    \caption{Description of the observation space.}
    \begin{tabular}{c|lcc}
        \toprule
        Env & Description & Range & Unit \\
        \midrule
        \multirow{8}{*}{\rotatebox[origin=c]{90}{MixedUse}} & Total power consumption & [0, 50] & kW \\
        & HVAC power consumption & [0, 50] & kW \\
        & Building power consumption & [0, 10] & kW \\
        & Outdoor relative humidity & [0, 100] & \%RH \\
        & Outdoor temperature & [-10, 40] & $^\circ$C \\
        & Zone 4 temperature & [10, 40] & $^\circ$C \\
        & Zone 5 temperature & [10, 40] & $^\circ$C \\
        & Avg. of other 6 Zones' temperature & [10, 40] & $^\circ$C \\
        \midrule
        \multirow{8}{*}{\rotatebox[origin=c]{90}{DataCenter}} & Total power consumption & [0, 200] & kW \\
        & HVAC power consumption & [0, 200] & kW \\
        & Building power consumption & [0, 200] & kW \\
        & Outdoor temperature & [-20, 50] & $^\circ$C \\
        & Outdoor relative humidity & [0, 100] & \%RH \\
        & West zone air temperature & [0, 50] & $^\circ$C \\
        & East zone air temperature & [0, 50] & $^\circ$C \\
        & Power utilization effectiveness & [1, 5] & - \\
        \bottomrule
    \end{tabular}%
    \label{tab_observation_desc}%
\end{table}%

\begin{table}[ht]
    \centering\small
    \caption{Description of the action space.}
    \begin{tabular}{c|lcc}
        \toprule
        Env & Description & Range & Unit \\
        \midrule
        \multirow{5}{*}{\rotatebox[origin=c]{90}{MixedUse}} & Zone setpoint temperature & [16, 26] & $^\circ$C \\
        & AHU 1 setpoint temperature & [10, 30] & $^\circ$C \\
        & AHU 2 setpoint temperature & [10, 30] & $^\circ$C \\
        & AHU 1 flow rate & [0, 1] & - \\
        & AHU 2 flow rate & [0, 1] & - \\
        \midrule
        \multirow{4}{*}{\rotatebox[origin=c]{90}{DataCenter}} & West zone setpoint temperature & [10, 40] & $^\circ$C \\
        & East zone setpoint temperature & [10, 40] & $^\circ$C \\
        & West zone mass flow rate & [1.75, 7] & kg/s \\
        & East zone mass flow rate & [1.75, 7] & kg/s \\
        \bottomrule
    \end{tabular}%
    \label{tab_action_desc}%
\end{table}%

\subsection{Environmental Variables}
    
    In this subsection, we detailed the experimental settings regarding two simulator environments and RL-based HVAC control algorithms. Important settings can be found in Table~\ref{tab_params}.

    \subsubsection{Observation Space}
    
    To make it clear, Table~\ref{tab_observation_desc} lists the variables composed of observation spaces for MixedUse and DataCenter. Although features including exogenous variables like weather conditions, indoor thermal state variables like temperatures, and HVAC-related variables like power consumption are all considered, it is still far from the complete description of the whole system. Hence, the final input variables is historical sequences of such an observation space with a fixed sequence length listed in Table~\ref{tab_params}.
    Moreover, to balance the units for different variables, a Min-Max normalization is imposed to ensure all values within unit intervals.

    \subsubsection{Action Space}
    
    All control variables of continuous action spaces in two environments are displayed in Table~\ref{tab_action_desc}.
    Here, we provide additional annotations for the action setting of MixedUse. We share the setpoint temperature of all indoor controllable zones with the same value, except for Zone 4. Since Zone 4 is the control room which has a unique thermal requirement, we fix the temperature setpoint as 18$^{\circ}C$.
    The reason for this operation is two-fold. First, except for zone 4, the other controllable indoor zones exhibit similar thermodynamic properties. Second, we empirically find that compared with the control of temperature and flow rate setpoints regarding two AHUs, the adjustment of indoor zone setpoint temperatures is much less influential.
    At last, similar to the observation space, all action values are normalized into the $[-1, 1]$ interval for stabilizing the training.

\begin{table*}[!ht]
    \centering\small
    \caption{Parameter Settings.}
    \begin{tabular}{l|cc}
        \toprule
        Environment & MixedUse & DataCenter \\
        \midrule
        Training / Testing Period & \multicolumn{2}{c}{1 year} \\ 
        \midrule
        Training weather files & \makecell{[GRC\_A\_Athens, \\GRC\_TC\_SkiathosAP, \\ GRC\_TC\_LarisaAP1, \\GRC\_TC\_Trikala]} & \makecell{[USA\_IL\_Chicago-OHare, \\USA\_CA\_San.Francisco, \\USA\_VA\_Sterling.Dulles, \\USA\_FL\_Tampa]}\\
        \midrule
        Testing weather file & GRC\_TC\_Lamia1 & CHN\_Hong.Kong.SAR \\
        \midrule
        Observation sequence length & 20 & 30 \\
        \midrule
        Reward weights $(\lambda_S, \lambda_T, \lambda_P)$ & (0.5, 0.1, $2 \times 10^{-5}$) & (0.5, 0.1, $10^{-5}$) \\
        \midrule
        Temperature target $T_C$ & 23.5 & 22 \\
        \midrule
        Temperature tolerance $[T_L^i, T_U^i]$ & [23, 24] & [21, 23] \\
        \midrule
        Discount factor $\gamma$ & \multicolumn{2}{c}{0.9} \\
        \midrule
        Training step & \multicolumn{2}{c}{10,000} \\
        \midrule
        Batch size & \multicolumn{2}{c}{256} \\
        \midrule
        Transformer - & \multicolumn{2}{c}{\multirow{2}{*}{(2, 4)}} \\
        (num\_blocks, num\_heads) & \\
        \midrule
        Feature size & \multicolumn{2}{c}{100} \\
        \midrule
        Hidden size & \multicolumn{2}{c}{200} \\
        \bottomrule
    \end{tabular}%
    \label{tab_params}%
\end{table*}%

    \subsubsection{Weather Files}

    As suggested in \cite{moriyama2018reinforcement}, training RL algorithms alternatively between different weather files helps the controller perform better with less power consumption. As such, we select different weather files in 4 locations for each environment, along with another one for evaluation.

\subsection{Baseline Method}
    
    We select the rule-based controller as the baseline control method. In general, the implementation follows the benchmark~\cite{scharnhorst2021energym}. As the exclusive difference, we remove the night mode for all rule-based controllers to align with the algorithm-based controllers.

\subsection{Reward Function}\label{sec_exp_rwd}

    Two prominent goals for HVAC control systems are thermal comfort for indoor zones as well as power consumption.
    Hence, the design of the reward function should also take into account both factors.
    To balance the penalty for temperature violations and the total energy cost, we follow the generic form described in the DataCenter~\cite{moriyama2018reinforcement} environment, which is exactly
    \begin{equation}
        r_t = r_T + \lambda_P \cdot r_P,
    \end{equation}
    where $\lambda_P$ is the energy weight that is responsible for scaling energy cost $r_P = - P_t$ and the temperature penalty $r_T$ is the weighted summation of Gaussian-shaped and trapezoid penalty parts, denoted by $r_{T_g}$ and $r_{T_t}$ respectively, for each zone $i=1, \cdots, z$, i.e.,
    \begin{equation}
        r_T = \sum_{i=1}^z \left(r^i_{T_g} + \lambda_{T} \cdot r^i_{T_t} \right).
    \end{equation}
    Here, the concrete forms of $ r^i_{T_g}$ and $r^i_{T_t}$ are
    \small{\begin{equation}\begin{split}
        r^i_{T_g} = & \exp \left( - \lambda_S \left( T_t^i-T_C^i \right)^2 \right), \\
        r^i_{T_t} = & \max \left( T_t^i - T_U^i, 0 \right) + \max \left( T_L^i - T_t^i, 0 \right),
    \end{split}\end{equation}}
    with $T_C^i $ as the target temperature, $\left[T_L^i, T_U^i \right]$ as the tolerance interval and $T_t^i$ as the real temperature for zone $i$, in addition to a Gaussian shape weight $\lambda_S$.
    Their practical values refer to the comparisons in \cite{biemann2021experimental} and are concluded in Table~\ref{tab_params}.

    \subsection{Performance Metrics}

    In order to compare the performance of RL algorithm-based controllers, we select the reward value as the main metric. However, as discussed in the previous subsection, thermal comfort and power consumption are two crucial ingredients of the reward value.
    Therefore, for the comparison of rule-based and algorithm-based controllers, we additionally consider two auxiliary metrics, i.e., the violation ratio of the indoor temperatures in terms of the tolerance intervals, as well as the average power consumption during the whole year.
    

    \subsection{Implementation Details}

    The implementation of the off-policy and offline RL algorithm-based controllers relies on the d3rlpy~\cite{seno2022d3rlpy} library. 
    In particular, after reforming the existing environments to store observation history sequences and fit for the Gymnasium APIs, we customize the encoder factories for the actor and the critic in d3rlpy.
    Moreover, all run time experiments were run with a single GeForce GTX 3090 GPU and an Intel Core i7-11700KF 3.60GHz CPU.
    The code is available at \url{https://github.com/cityujun/hvac_offline_rl}.

    \subsection{Research Questions}

    Generally, we empirically investigate the following research questions. 
    \begin{enumerate}
        \item[\textbf{RQ1}] Do the normal off-policy RL algorithms manage an effective control for the offline scenarios?
        
        \item[\textbf{RQ2}] Does the integration of the observation history modeling achieve better outcomes compared with normal methods?

        \item[\textbf{RQ3}] How does the quality of offline datasets influence the performance of offline RL algorithms?

        \item[\textbf{RQ4}] How does the scale of offline datasets affect the performance of offline RL algorithms?

        \item[\textbf{RQ5}] How is the sensitivity of the hyperparameters of algorithm-based controllers? For instance, if the integration of the observation history modeling really works, how does the performance gain change as the length of observation history increases?
    \end{enumerate}

\section{Results and Discussions}\label{sec_exp_results}
In this section, we present the results and in-depth analysis of the above research questions in turn.

\subsection{Comparisons on Off-policy and Offline Algorithms}\label{sec_exp_rq1}

    To answer RQ1, we design two scenarios where offline data is generated quite differently, denoted by \textbf{Final Buffer} and \textbf{Trained} respectively.
    Then, through the comparisons of off-policy and offline RL controllers for HVAC systems, we aim to validate the necessity of offline RL algorithms for purely data-driven HVAC controllers. 
    \begin{itemize}\small
        \item \textbf{Scenario 1 (Final Buffer)} We first train an off-policy algorithm-based (e.g., SAC) controller from scratch for 3 million steps. During this process, sufficient exploration is allowed with $\mathcal{N}(0,0.1)$ Gaussian noises at each step. We store all rollout transitions that the controller interacts with the environment.

        \item \textbf{Scenario 2 (Trained)} A well-trained off-policy algorithm-based (e.g., SAC) controller is employed to interact with the environment for 3 million steps, and all transitions are stored. In this process, the same noise is imposed uniformly distributed at only 10\% of the total steps.
    \end{itemize}
    As such, in the first scenario, due to sufficient exploration, the collected dataset contains a diverse set of states and actions, leading to a considerable coverage of the entire state-action space. For the second scenario, the trained SAC agent performs as an expert. Meanwhile, the collected offline data is of higher quality but has limited state-action space coverage.

    \begin{figure*}[ht]
        \centering
        \includegraphics[width=1.\linewidth]{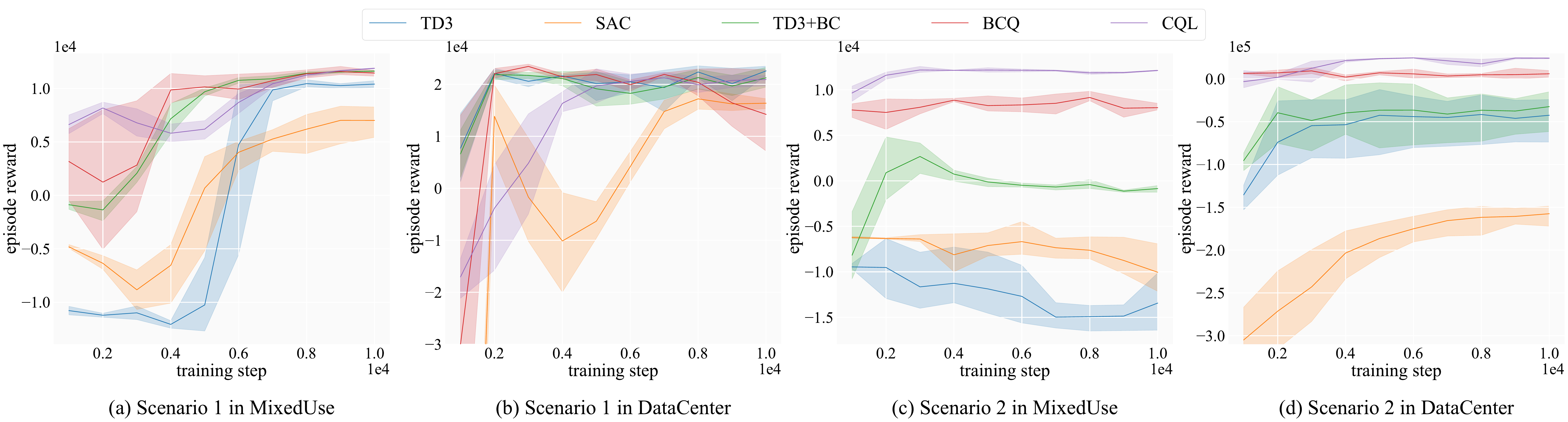}
        \caption{Comparisons of off-policy and offline RL algorithms in two different scenarios.}
        \label{fig_secnario_comparison}
    \end{figure*}

    From the results in Figure~\ref{fig_secnario_comparison}, it is clearly observed that with sufficient coverage (i.e., Scenario 1), both off-policy and offline RL algorithm-based HVAC controllers can achieve great performance finally in MixedUse and DataCenter. However, offline RL algorithms (CQL, BCQ and TD3+BC) still display obvious advantages in learning efficiency. Beneficial to the existence of regularization terms, they are not allowed to learn policies deviating from the behavioral ones contained in offline data.
    
    More importantly, when it comes to high-quality data with insufficient coverage (i.e., Scenario 2), the state-of-the-art off-policy algorithm-based controllers (TD3 and SAC) perform quite worse, which aligns with our expectations. Without online interactions with the environment, their actors are freely optimized against the critic, leading to policy evaluation and policy improvement that are both out of control.
    By contrast, HVAC controllers relying on offline RL algorithms (CQL, BCQ, and TD3+BC) still exhibit a compelling performance, which explicitly confirms the significance of mitigating extrapolation errors during training controllers in terms of offline data.
    
    Finally, among all offline RL algorithms, CQL-based HVAC controllers tend to achieve the best reward performance. Therefore, we select CQL as the backbone offline RL algorithm for the remaining experiments. The reason that TD3+BC performs relatively worse (even when we impose a higher penalty weight to encourage imitation), might be the $L_2$ regularization is not compatible with our environmental settings.

    \begin{table*}[t] 
        \centering\small
        \caption{The performance of rule-based and off-policy algorithm-based HVAC controller.}\label{tab_expert_perf}
        \setlength{\tabcolsep}{4pt}
        \begin{threeparttable}
            \begin{tabular}{c|l|ccc|ccc|ccc}
                \toprule
                \multicolumn{1}{c}{\multirow{2}{*}{}} & \multicolumn{1}{c}{\multirow{2}{*}{\makecell{Weather \\ Location}}}  & \multicolumn{3}{c}{Rule-based} & \multicolumn{3}{c}{Off-Policy w/o history}  & \multicolumn{3}{c}{Off-Policy w/ history} \\
                \cmidrule(lr){3-5} \cmidrule(lr){6-8} \cmidrule(lr){9-11} \multicolumn{1}{c}{} & \multicolumn{1}{c}{} & A.R. & T.V. & \multicolumn{1}{c}{A.P.} & A.R. & T.V. & \multicolumn{1}{c}{A.P.} & A.R. & T.V. & A.P. \\
                \midrule
                \multirow{5}{*}{\rotatebox[origin=c]{90}{MixedUse}} & A\_Athens & 0.33 & 24.1\% & 13.27 & 0.54 & 34.1\% & 15.56 & \textbf{0.60} & \textbf{18.3\%} & \textbf{11.81} \\
                & TC\_SkiathosAP & 0.30 & 26.5\% & 14.20 & 0.42 & 41.0\% & 16.38 & \textbf{0.57} & \textbf{19.7\%} & \textbf{12.76} \\
                & TC\_LarisaAP1 & 0.17 & 26.3\% & 16.62 & 0.31 & 40.4\% & 18.53 & \textbf{0.44} & \textbf{20.4\%} & \textbf{14.65} \\
                & TC\_Trikala & 0.00 & 38.0\% & 19.79 & 0.07 & 48.0\% & 21.24 & \textbf{0.14} & \textbf{34.1\%} & \textbf{17.91} \\
                \cmidrule(lr){2-11} & TC\_Lamia1 & 0.07 & 35.0\% & 17.50 & 0.21 & 46.0\% & 18.99 & \textbf{0.33} & \textbf{28.0\%} & \textbf{15.38} \\
                \midrule
                \multirow{5}{*}{\rotatebox[origin=c]{90}{DataCenter}} & IL & 0.13 & 42.5\% & 117.4 & 0.29 & 29.9\% & \textbf{107.1} & \textbf{0.60} & \textbf{2.62\%} & 126.4 \\
                & CA & 0.18 & 41.1\% & 113.9 & 0.56 & 15.9\% & \textbf{108.0} & \textbf{0.67} & \textbf{2.60\%} & 121.0 \\
                & VA & 0.10 & 43.9\% & 119.2 & 0.34 & 26.8\% & \textbf{108.3} & \textbf{0.59} & \textbf{2.67\%} & 128.0 \\
                & FL & -0.02 & 32.6\% & 135.4 & \textbf{0.53} & 7.03\% & \textbf{127.0} & 0.47 & \textbf{2.41\%} & 142.5 \\
                \cmidrule(lr){2-11} & HK & -0.02 & 30.9\% & 136.8 & \textbf{0.57} & 4.13\% & \textbf{127.6} & 0.47 & \textbf{2.36\%} & 143.9 \\
                \bottomrule
            \end{tabular}%
            \begin{tablenotes}
                \item A.R. means the average reward of each step within an episode, T.V. means the averaged violation ratio of the indoor zone temperatures in terms of the tolerance interval, and A.P. means the average value of total power consumed (kW). The bold numbers indicate the best performance.
            \end{tablenotes}
        \end{threeparttable}
    \end{table*}%

    \begin{figure*}[!ht]
        \centering
        \begin{subfigure}[ht]{1.\linewidth}
            \includegraphics[width=1.\textwidth]{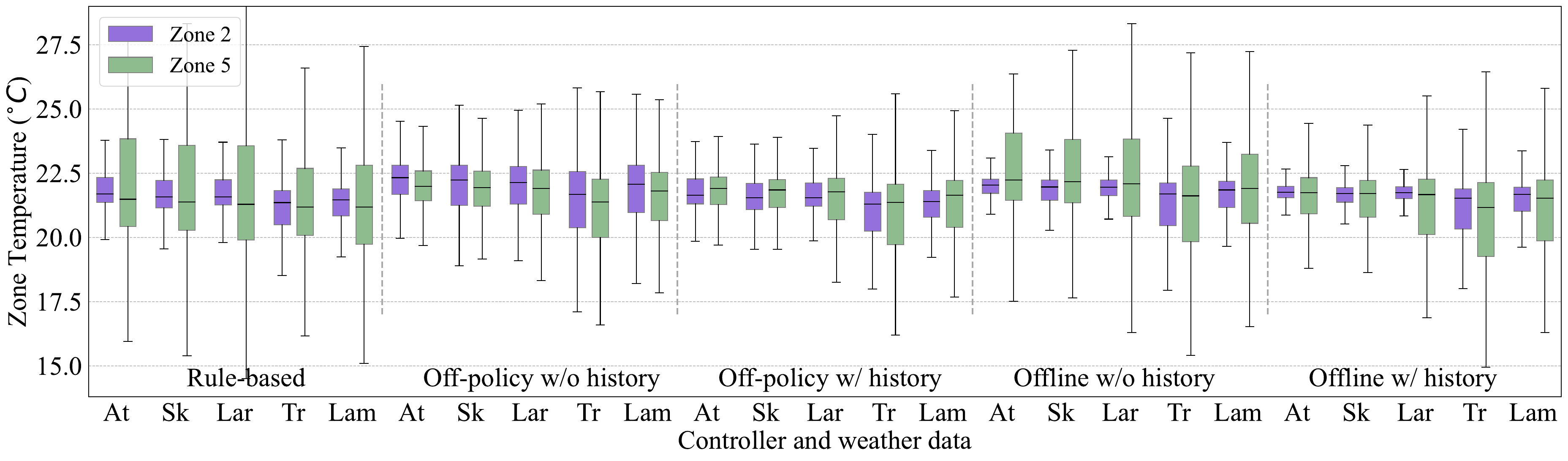}
            \subcaption{MixedUse}
            \label{fig_tempbox_mixeduse}
        \end{subfigure}
        \hfill
        \begin{subfigure}[ht]{1.\linewidth}
            \includegraphics[width=1.\textwidth]{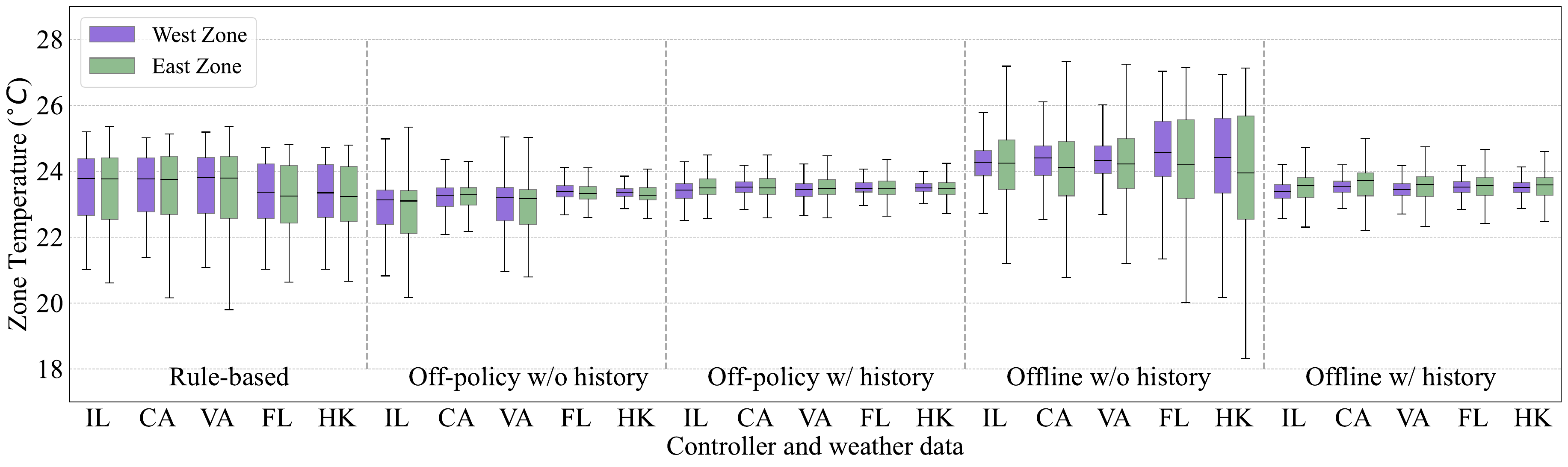}
            \subcaption{DataCenter}
            \label{fig_tempbox_datacenter}
        \end{subfigure}
        \caption{The box plots of indoor temperatures regarding various controllers. Basically, applying RL algorithms as well as the observation history modeling can achieve the most steady indoor temperatures.}
    \end{figure*}

\subsection{Effect of Observation History Modeling}\label{sec_exp_rq2}

    To investigate the effect of our innovative modeling for observation history sequences, we implement both off-policy and offline algorithm-based controllers with and without such a component.
    
    In Table~\ref{tab_expert_perf}, we report the main performance metrics considered in terms of the off-policy algorithm-based controllers. Note that CQL-based controllers trained from offline data generated by SAC or TD3 policies display results similar to those of SAC or TD3.
    First of all, compared with the baseline rule-based controllers, deploying RL algorithm-based controllers can definitely result in more rewarding actions.
    However, in MixedUse, simply applying such controllers without considering observation histories might not ultimately lead to satisfactory outcomes since more power is consumed with higher violation ratios regarding the temperature tolerance intervals.

    Fortunately, by incorporating observation history sequences, RL algorithm-based HVAC controllers generally outperform other controllers in gaining better rewards and lower temperature violations in both facilities. For example, in DataCenter, when compared with the rule-based controller, the average violation ratio of indoor temperatures drops from 30.9\% to 2.36\%, while experiencing only a 5\% increase in power consumption in terms of HK weather file.
    The reason why the decline in thermal comfort conditions results in extra energy usage may be the significant climatic differences between training and evaluation locations.
    As a comparison, for the MixedUse facility where all weather files associate locations in Greece, not only does the temperature violation enjoy an absolute decrease of 7\%, but the electricity usage is reduced by about 12.1\% in terms of the evaluation weather file.
    
    In fact, we further find that these policies can control indoor zone temperatures more steadily, which is beneficial from the touch of observation tendencies.
    Specifically, we plot the yearly distribution of indoor temperatures in terms of two representative zones (i.e., Zone 2 and Zone 5) of MixedUse and west/east zones of DataCenter in Figure~\ref{fig_tempbox_mixeduse} and Figure~\ref{fig_tempbox_datacenter} separately.
    It is clearly observed that with the onservation history module, the indoor temperatures enjoy a lower fluctuation, which helps improve the occupants' thermal comfort, especially in DataCenter.

    Consequently, we will include this module by default in the expert control algorithm for subsequent experiments.



\begin{figure*}[ht]
        \centering
        \begin{subfigure}[ht]{0.325\linewidth}
            \includegraphics[width=1.\textwidth]{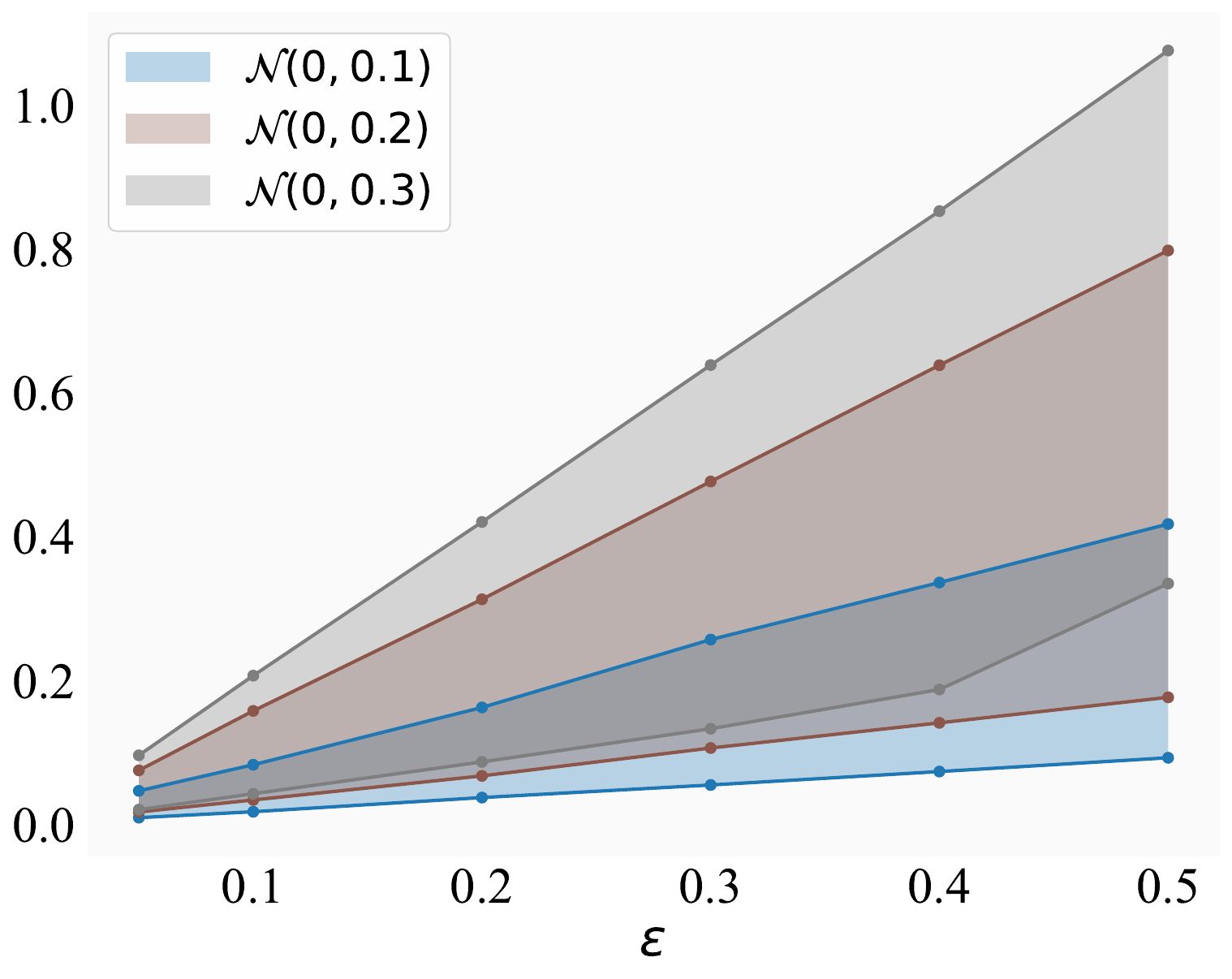}
            \subcaption{Range of $\delta_{\tau}$ in MixedUse}
        \end{subfigure}
        \hfill
        \begin{subfigure}[ht]{0.325\linewidth}
            \includegraphics[width=1.\textwidth]{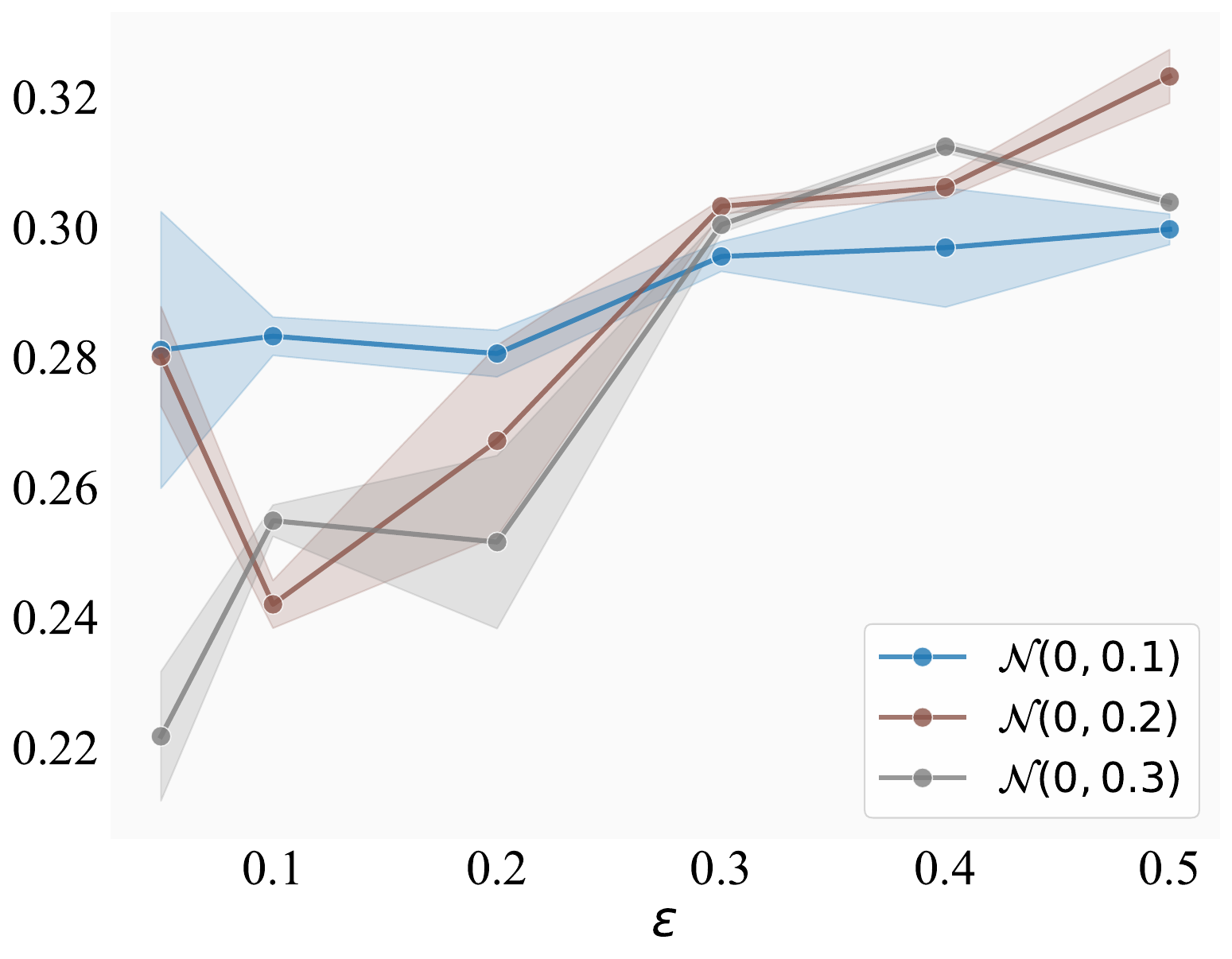}
            \subcaption{Step reward in MixedUse}
        \end{subfigure}
        \hfill
        \begin{subfigure}[ht]{0.325\linewidth}
            \includegraphics[width=1.\textwidth]{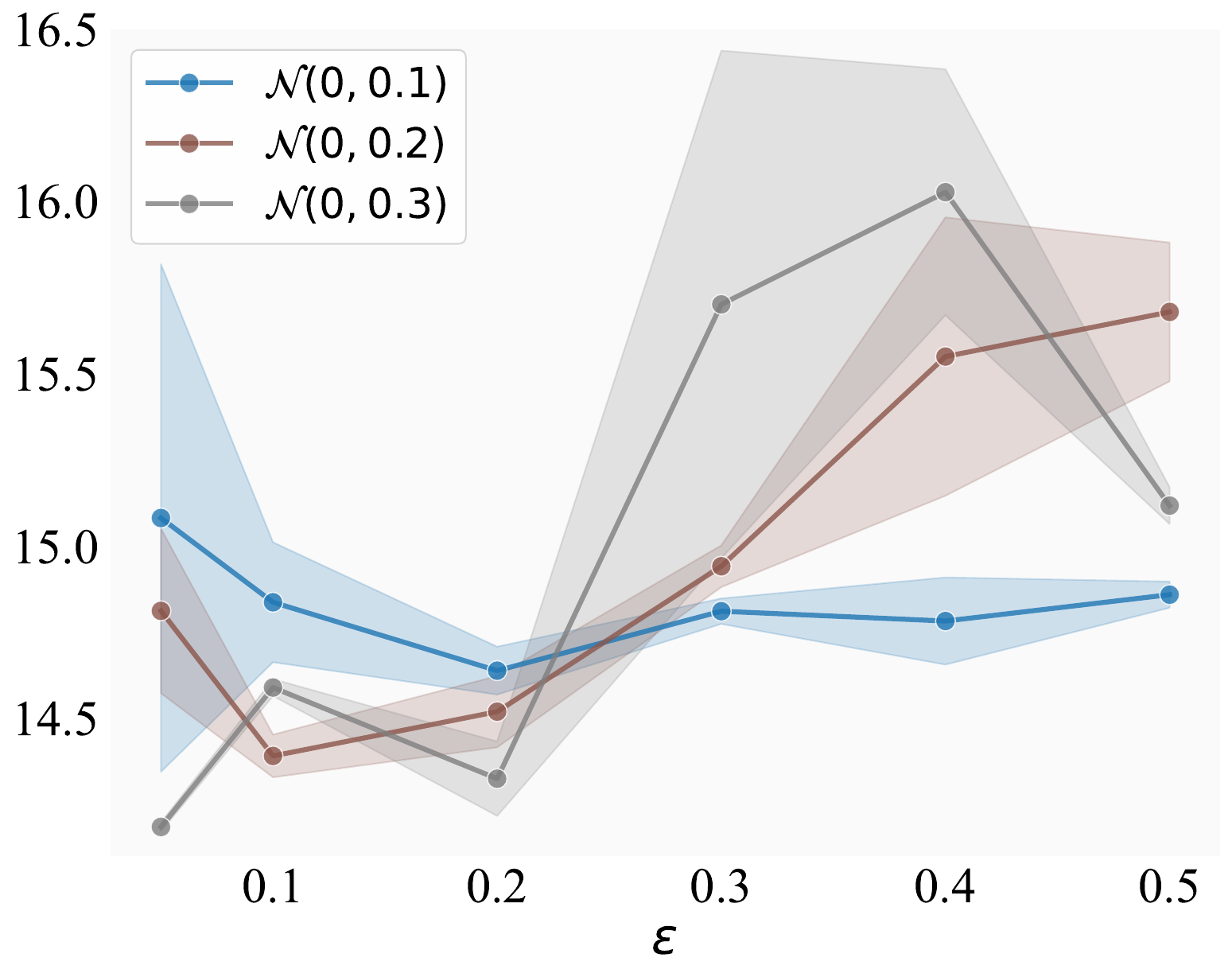}
            \subcaption{Average total power (kW) in MixedUse}
        \end{subfigure}
        \hfill
        \begin{subfigure}[ht]{0.325\linewidth}
            \includegraphics[width=1.\textwidth]{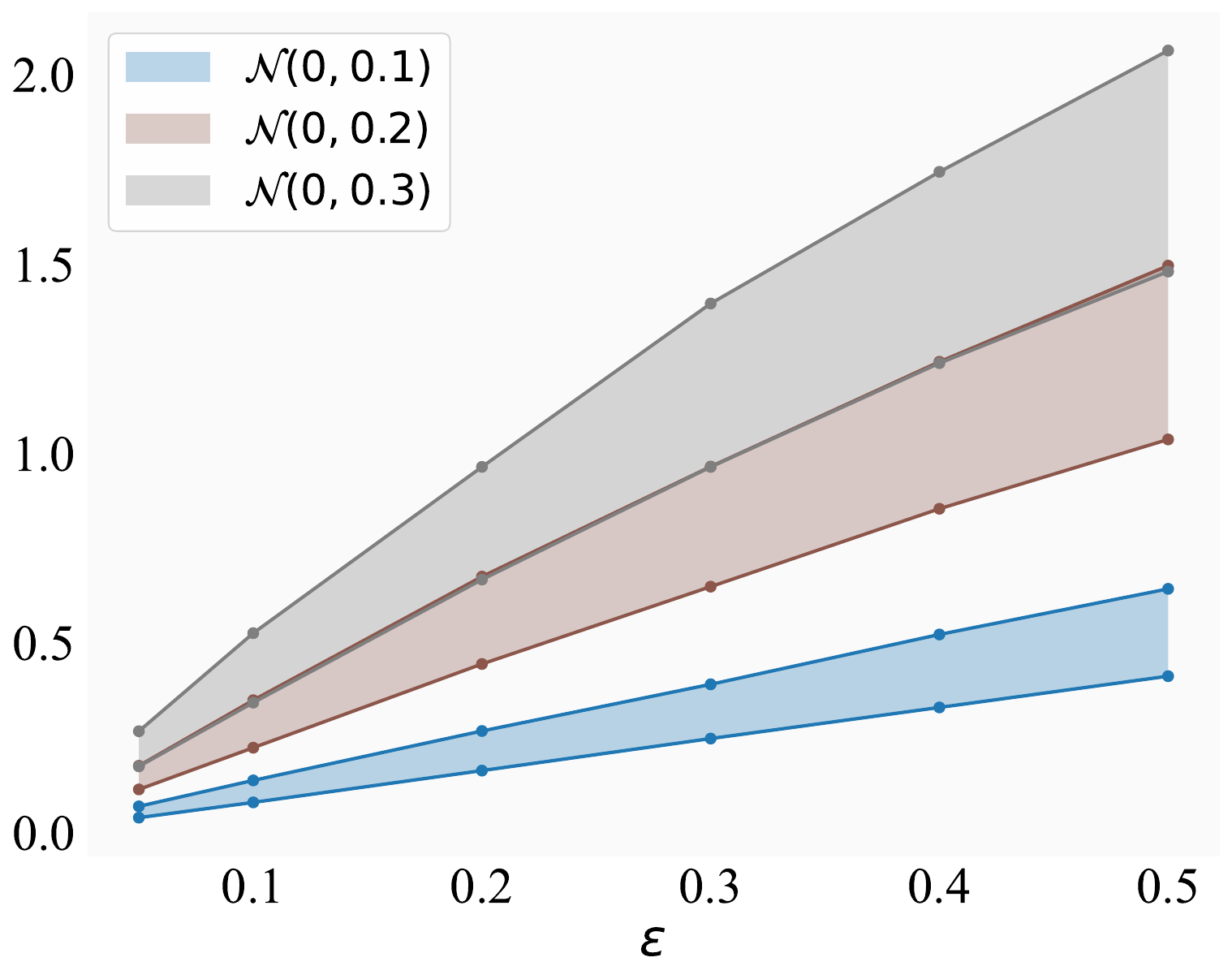}
            \subcaption{Range of $\delta_{\tau}$ in DataCenter}
        \end{subfigure}
        \hfill
        \begin{subfigure}[ht]{0.325\linewidth}
            \includegraphics[width=1.\textwidth]{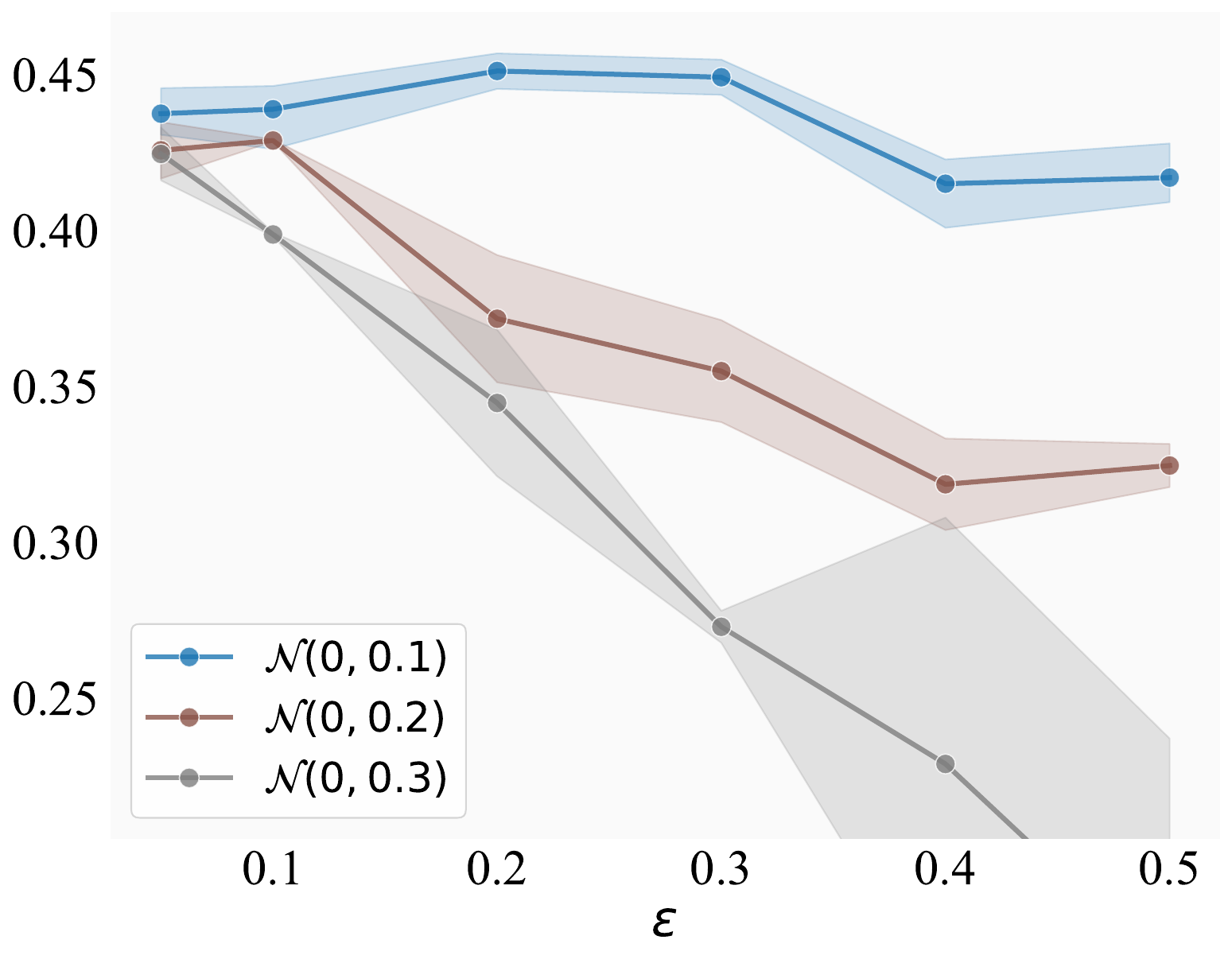}
            \subcaption{Step reward in DataCenter}
        \end{subfigure}
        \hfill
        \begin{subfigure}[ht]{0.325\linewidth}
            \includegraphics[width=1.\textwidth]{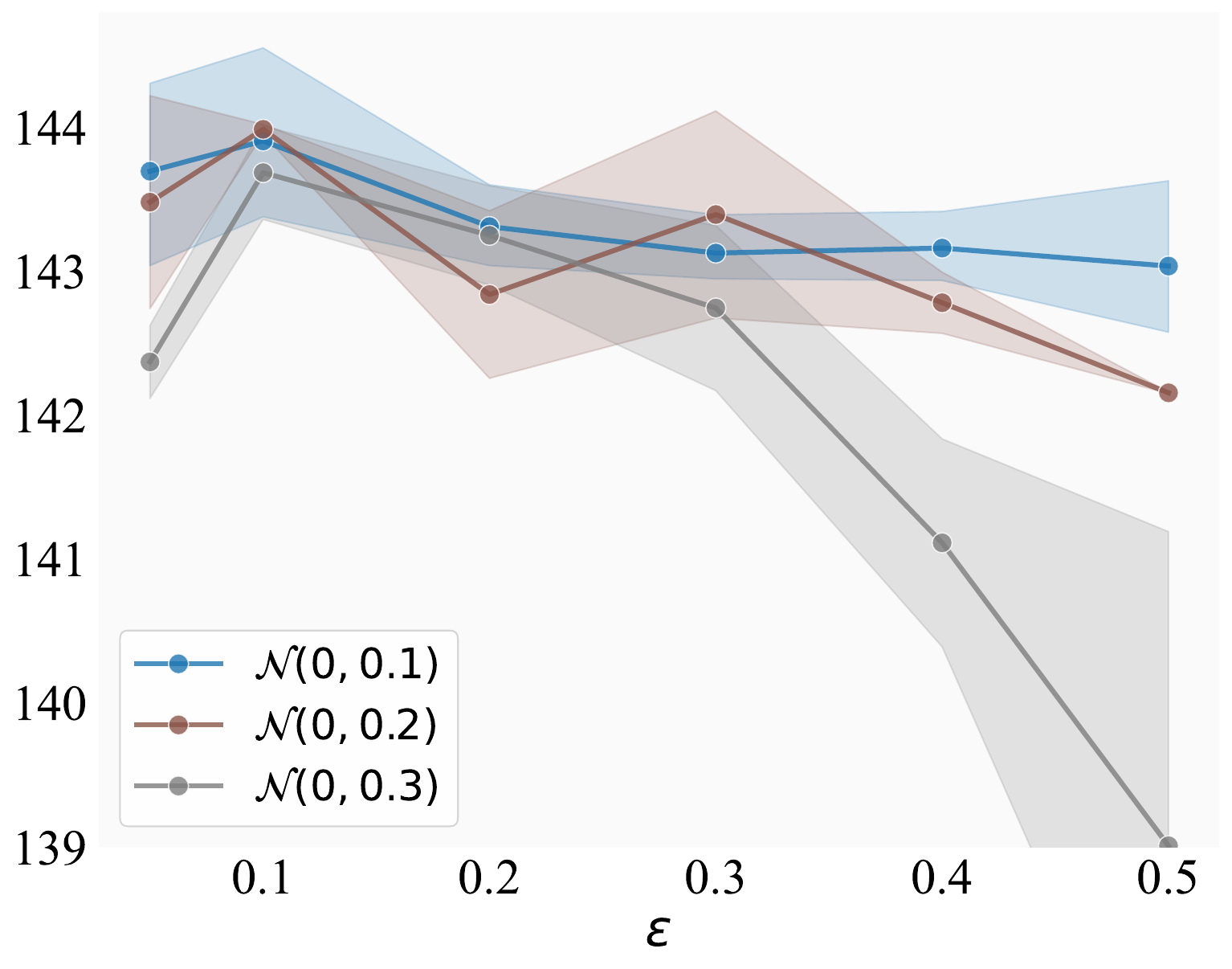}
            \subcaption{Average total power (kW) in DataCenter}
        \end{subfigure}
        \caption{Through applying different exploration ratio $\epsilon$ and normally distributed noises with different variance, we investigate the quality of generated offline data by means of defined regret ratio $\delta_{\tau}$, the step reward in an episode that the offline trained algorithm-based controller can gain, along with the average power totally consumed.}
        \label{fig_quality}
    \end{figure*}
    
\subsection{Investigation on the Quality of Offline Data}\label{sec_exp_rq3}

    Before addressing Research Question 3 (RQ3), a description of the quality of the collected offline data should be formally provided.
    Drawing on the regret concept from online algorithms, we assess the quality of trajectories with fixed-horizon in $\mathcal{D}$ by comparing the actual trajectory reward $r_{\tau}$ with the (estimated) optimal value $r_{opt}$ from a well-trained HVAC controller.
    To facilitate comparisons across various building environments, we use a normalized ratio value 
    \begin{equation}
        \delta_{\tau} = \frac{r_{opt} - r_{\tau}}{r_{opt}}
    \end{equation}
    to standardize the value scale. Intuitively, a smaller $\delta_{\tau}$ close to 0 indicates a more successful trajectory. Thus, the entire quality of the offline data $\mathcal{D}$ can be characterized by the distribution properties of $\delta_{\tau}$ values, such as $\left[\inf_{\tau} \delta_{\tau}, \sup_{\tau} \delta_{\tau}\right]$.

    To empirically obtain datasets with varying $\delta_{\tau}$, we introduce an exploration probability $\epsilon$. By uniformly adding Gaussian noise to the outputs of expert policies at each timestep, we generate suboptimal datasets without drastically altering the state's marginal distribution. Therefore, through the control of $\epsilon$, we actually manipulate the deviation degree from expert policies, thus changing the distribution of collected trajectories.

    The quantitative results for different $\epsilon$ values and Gaussian noises across two buildings are reported in Figure~\ref{fig_quality}.
    The first column of Figure~\ref{fig_quality} displays the range of $\delta_{\tau}$ in the generated datasets.
    It is important to note that weather files have a greater impact on $\delta_{\tau}$ than the randomness introduced by exploration probability $\epsilon$.
    In general, as $\epsilon$ values and Gaussian noise variances increase, the regret ratios of the trajectories rise almost linearly, exhibiting greater variances.
    Comparatively, regret ratios in DataCenter grow faster than in MixedUse, with less fluctuation across different weather files.
    The longer horizon setting in DataCenter (6 runtime control per hour) may explain the significantly higher regret ratios.

    Secondly, we separately train CQL-based controllers using generated offline datasets. To precisely reflect on the abilities of these controllers, an online evaluation process is followed to assess the average step reward in the whole episode.
    Ignoring the offline policy selection strategy~\cite{paine2020hyperparameter}, our goal is to explore the ground-truth assessment before deployment. Therefore, all training epochs are evaluated to obtain the best outcome.
    The results are summarized in the middle column of Figure~\ref{fig_quality}.
    In DataCenter, with the sharply decreased quality of datasets, the optimal reward performance is reached when $\epsilon=0.2$ for $\mathcal{N}(0, 0.1)$ noise, and $\epsilon=0.1$ for $\mathcal{N}(0, 0.2)$ noise.
    However, due to the increased lower bounds of $\delta_{\tau}$ for $\mathcal{N}(0, 0.3)$ noises, the initial $\epsilon$ value (i.e., 0.05) is the best choice.
    A similar trend is observed in MixedUse.
    Given the slower increase of $\delta_{\tau}$'s upper and lower bounds compared to DataCenter, the optimal reward occurs when $\epsilon$ ranges from 0.3 to 0.5 for different Gaussian noises.

    Last but not least, we also present the yearly average electricity expenses for the entire facility and the HVAC system in the third column of Figure~\ref{fig_quality}. Roughly, these expenses follow the trend of the reward values as the $\epsilon$ value increases. This suggests that selected reward weights $(\lambda_S, \lambda_T, \lambda_P)$ can appropriately balance the occupants' thermal comfort and electricity saving requirement.
    Another observation is that the energy consumption of CQL-based controllers is similar to that of off-policy SAC-based controllers, indicating their energy efficiency in MixedUse compared to the baseline of 17.50 kW from rule-based controllers.
    
    Surprisingly, a high-level conclusion from Figure~\ref{fig_quality} reveals that offline datasets with trajectories having small $\delta_{\tau}$ values near 0 are less suitable for training RL algorithm-based controllers, which might conflict with our intuition.
    In contrast, a mixture of suboptimal trajectories is beneficial to the reward performance in terms of offline-trained HVAC policies.
    As discussed in \cite{kumar2022should}, the power of offline RL algorithms lies in their ability to stitch trajectories from suboptimal data, unlike supervised methods that depend on the most frequent actions. Therefore, trajectories with $\delta_{\tau}$ values properly greater than 0 can balance the diversity of transitions and the ``quality'' of gathered datasets.

\begin{figure}[ht]
    \centering
    \begin{subfigure}[ht]{0.48\textwidth}
        \includegraphics[width=1.\textwidth]{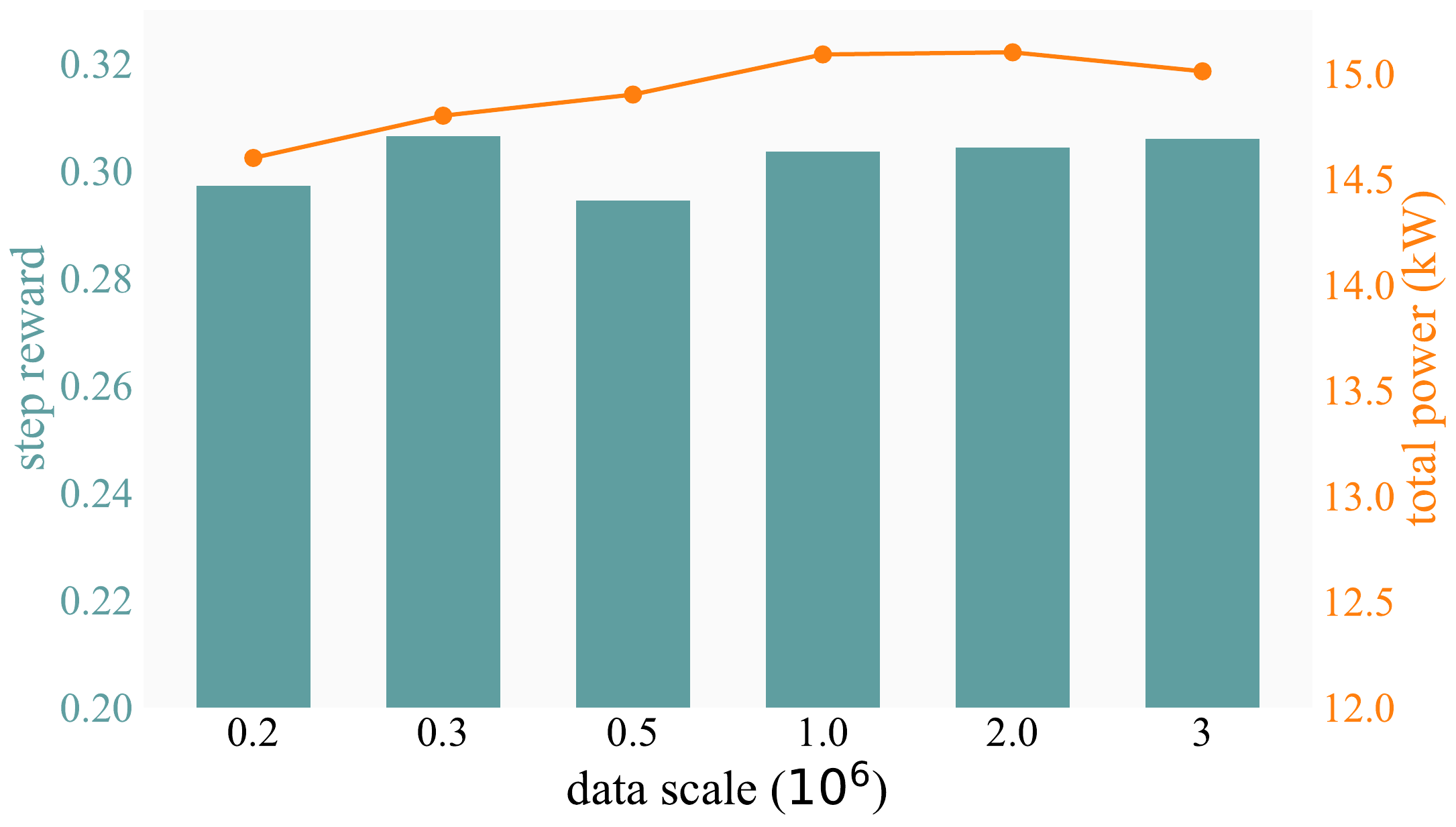}
        \subcaption{MixedUse}
    \end{subfigure}
    \hfill
    \begin{subfigure}[ht]{0.48\textwidth}
        \includegraphics[width=1.\textwidth]{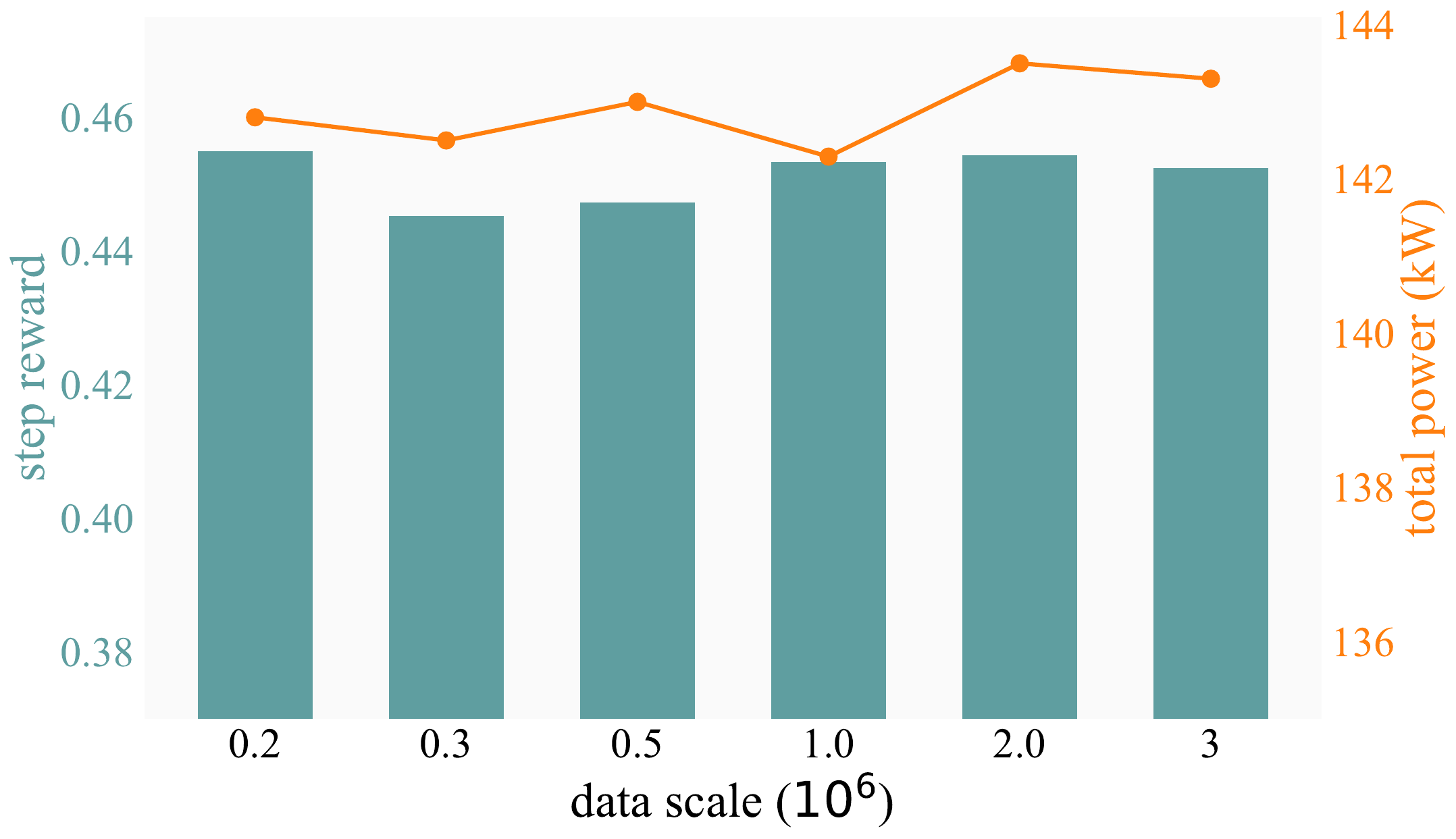}
        \subcaption{DataCenter}
    \end{subfigure}
    \caption{Comparisons of the scale of gathered datasets for offline training of CQL-based HVAC controllers.}
    \label{fig_quantity}
\end{figure}

\subsection{Investigation on the Quantity of Offline Data}\label{sec_exp_rq4}
    
    Next, we examine another key factor that could remarkably influence the deployment of algorithm-based controllers, i.e., the quantity demand for offline datasets.
    Off-policy training manner is known to require a large number of transitions. For example, as detailed in Section~\ref{sec_exp_rq1}, the scale of $10^6$ is necessary for convergence when we train the SAC-based controller in the final buffer scenario.
    Therefore, we reduce the size of the gathered datasets incrementally to observe performance degradation while keeping the exploration probability $\epsilon$ as well as the variance of Gaussian noises constant. Specifically, we set $(\epsilon, \sigma)$—with $\sigma$ representing $\mathcal{N}(0, \sigma)$—to (0.5, 0.2) for MixedUse and (0.2, 0.1) for DataCenter.

    The performance of step reward values in addition to the energy consumption is shown in Figure~\ref{fig_quantity}.
    It can be found that reducing the size of offline datasets from $10^6$ to $10^5$ would not evidently affect the control performance of HVAC controllers.
    Considering that an episode includes more than $10^4$ transitions over a year, only a few episodes are necessary for algorithm-based controllers to achieve satisfactory performance. 
    As illustrated by the orange dashed lines, the energy demand exhibits a relatively steady tendency for policies with similar reward performance, aligning with the findings in Section~\ref{sec_exp_rq3}.

\begin{figure}[ht]
    \centering
    \begin{subfigure}[ht]{0.48\linewidth}
        \includegraphics[width=1.\textwidth]{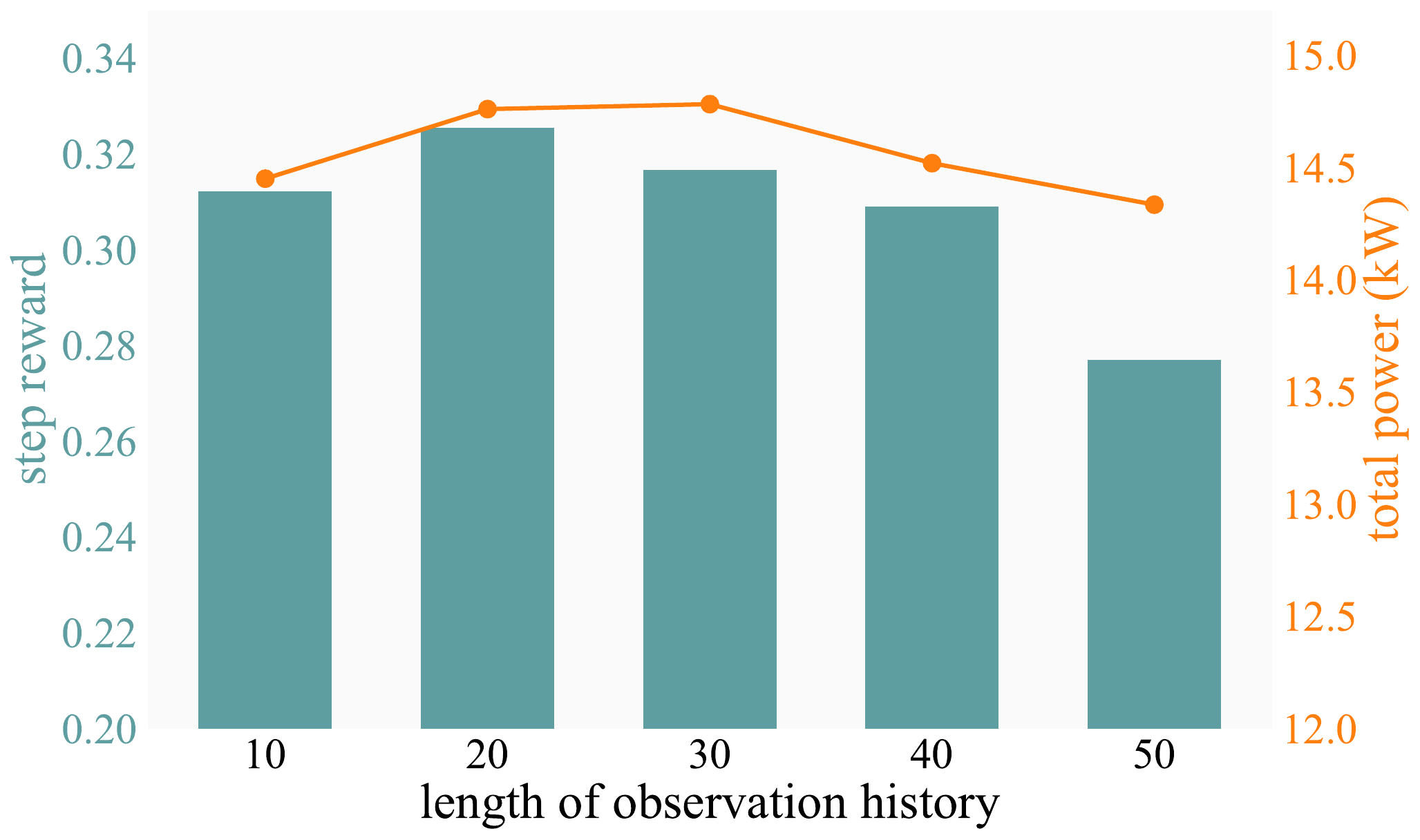}
        \subcaption{MixedUse}
    \end{subfigure}
    \hfill
    \begin{subfigure}[ht]{0.48\linewidth}
        \includegraphics[width=1.\textwidth]{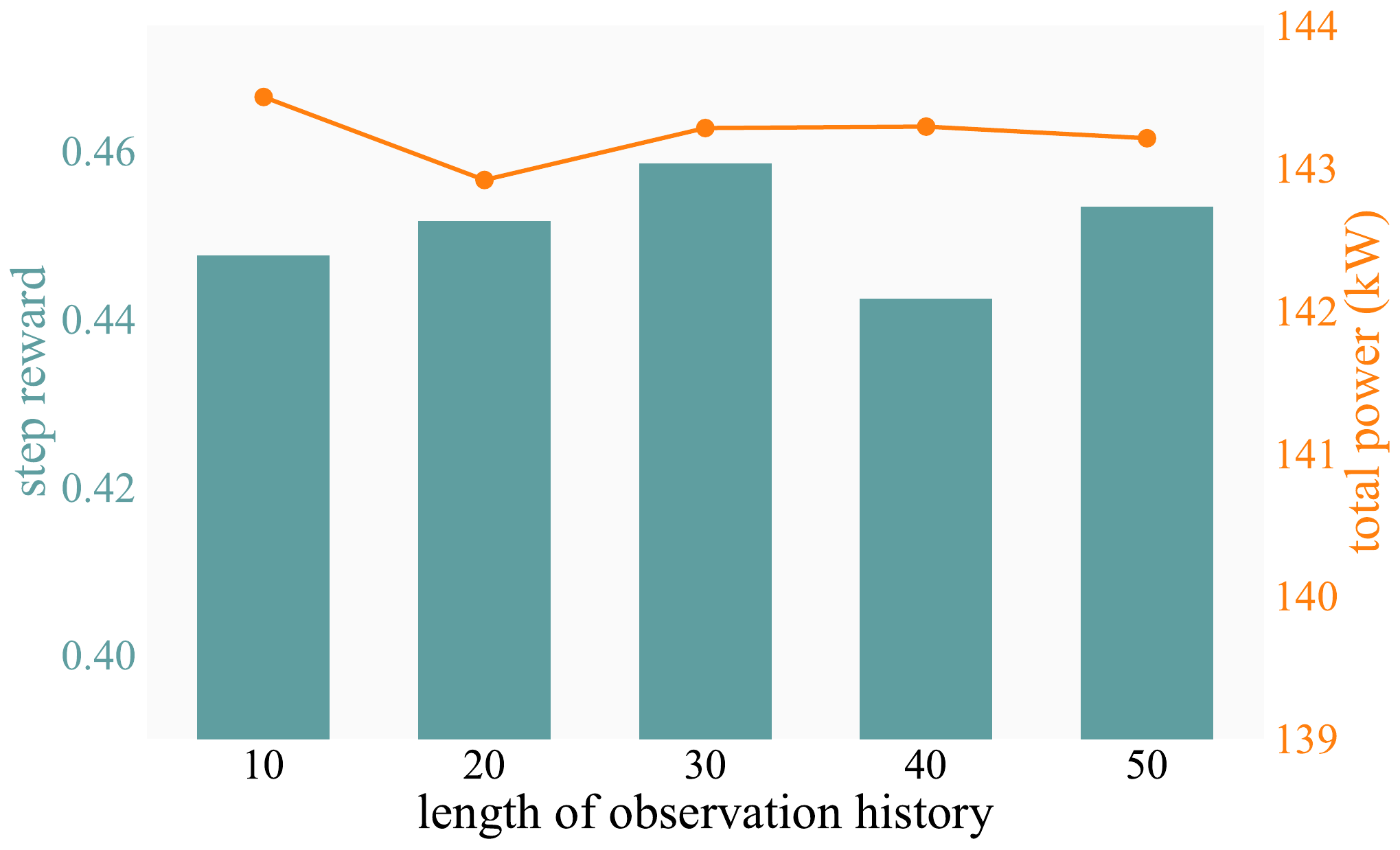}
        \subcaption{DataCenter}
    \end{subfigure}
    \caption{Comparisons of the sequence length of observation history for offline training of CQL-based HVAC controllers.}
    \label{fig_sequence_length}
\end{figure}

\subsection{Investigation on the Length of Observation History}\label{sec_exp_rq5}

    We also observed that certain hyperparameters significantly impact the performance of RL algorithm-based controllers. Specifically, we highlight the crucial role of sequence length in modeling the observation histories, which actually induces a trade-off between model capacity and efficiency.
    In other words, extending the observation horizon offers more historical information for decision-making at the cost of a higher computational requirement.

     As in Section~\ref{sec_exp_rq4}, we demonstrate the performance in terms of step reward values and energy consumption in Figure~\ref{fig_sequence_length}.
     We found that increasing the sequence length of the observation horizon up to 50 does not significantly benefit the reward performance as anticipated.
     Therefore, we deduce that the chosen sequence lengths (20 for MixedUse and 30 for DataCenter) sufficiently capture past trends. Conversely, longer sequence lengths may introduce unnecessary noise into the current HVAC control decisions.

\section{Conclusion}\label{sec_conclusion}

Algorithm-based HVAC controllers are crucial for smart building HVAC systems, particularly those using reinforcement learning methods.
This paper systematically investigates offline RL algorithm-based HVAC controllers to explore pure data-driven solutions without relying on expert experience.
Firstly, we concentrate on the end-to-end learning framework, integrating the benefits of observation history modeling.
Secondly, we quantitatively analyze the relationship between the collected datasets and the control policies trained offline.
As a result, our findings first reveal that in the field of building HVAC systems, datasets with suitable randomness on expert decisions are more favorable for training a data-driven control strategy, along with a much lower size demand for the datasets compared with off-policy methods.

In the future, an interesting direction is to understand the impact of a mixture of datasets with various defined properties on the resulting control strategies. In addition, it is also crucial to explore a better integration manner of the observation history module. For example, introducing the contrastive term in order to put an emphasis on the transition with higher rewards could further enhance the representation ability of sequential models as well as the training efficiency of HVAC controllers. 






\bibliography{references}

\end{document}